\documentclass[iicol,sn-mathphys-num]{sn-jnl}

\usepackage{graphicx}%
\usepackage{multirow}%
\usepackage{amsmath,amssymb,amsfonts}%
\usepackage{amsthm}%
\usepackage{mathrsfs}%
\usepackage[title]{appendix}%
\usepackage{xcolor}%
\usepackage{textcomp}%
\usepackage{manyfoot}%
\usepackage{booktabs}%
\usepackage{listings}%

\theoremstyle{thmstyleone}%
\theoremstyle{thmstyletwo}%

\theoremstyle{thmstylethree}%

\usepackage{tabularx}
\newcolumntype{C}{>{\centering\arraybackslash}X}
\newcolumntype{M}[1]{>{\centering\arraybackslash}p{#1}}
\usepackage{booktabs}
\usepackage{makecell}
\usepackage{adjustbox}
\usepackage{multirow}
\usepackage[table]{xcolor}
\definecolor{dino}{RGB}{249,231,227}
\definecolor{dino_text}{RGB}{209,154,128}

\usepackage{setspace}
\usepackage[norelsize,ruled]{algorithm2e}

\definecolor{commentcolor}{RGB}{110,154,155}   
\newcommand{\PyComment}[1]{\ttfamily\textcolor{commentcolor}{\# #1}}  
\newcommand{\PyCode}[1]{\ttfamily\textcolor{black}{#1}} 
\usepackage{ragged2e}
\begin{document}

\title[Multiple Instance Learning Framework with Masked Hard Instance Mining for Gigapixel Histopathology Image Analysis]{Multiple Instance Learning Framework with Masked Hard Instance Mining for Gigapixel Histopathology Image Analysis}

\author[1]{\fnm{Wenhao} \sur{Tang}}\email{whtang@cqu.edu.cn}
\author*[1]{\fnm{Sheng} \sur{Huang}}\email{huangsheng@cqu.edu.cn}
\author[1]{\fnm{Heng} \sur{Fang}}\email{fangheng@cqu.edu.cn}
\author[2]{\fnm{Fengtao} \sur{Zhou}}\email{fzhouaf@connect.ust.hk}
\author[3]{\fnm{Bo} \sur{Liu}}\email{kfliubo@gmail.com}
\author[4]{\fnm{Qingshan} \sur{Liu}}\email{qsliu@njupt.edu.cn}

\affil*[1]{\orgdiv{School of Big Data \& Software Engineering}, \orgname{Chongqing University}, \orgaddress{\state{Chongqing}, \country{China}}}

\affil[2]{\orgdiv{CS}, \orgname{Hong Kong University of Science and Technology}, \orgaddress{\state{Hong Kong}, \country{China}}}

\affil[3]{\orgdiv{CS}, \orgname{Hefei University of Technology}, \orgaddress{\state{Hefei}, \country{China}}}

\affil[4]{\orgdiv{CS}, \orgname{Nanjing University of Posts and Telecommunications}, \orgaddress{\state{Nanjing}, \country{China}}}


\abstract{Digitizing pathological images into gigapixel Whole Slide Images (WSIs) has opened new avenues for Computational Pathology (CPath).
As positive tissue comprises only a small fraction of gigapixel WSIs, existing Multiple Instance Learning (MIL) methods typically focus on identifying salient instances via attention mechanisms.
However, this leads to a bias towards easy-to-classify instances while neglecting challenging ones. 
Recent studies have shown that hard examples are crucial for accurately modeling discriminative boundaries.
Applying such an idea at the instance level, 
we elaborate a novel MIL framework with masked hard instance mining (MHIM-MIL),
which utilizes a Siamese structure with a consistency constraint to explore the hard instances.
Using a class-aware instance probability, MHIM-MIL employs a momentum teacher to mask salient instances and implicitly mine hard instances for training the student model.
To obtain diverse, non-redundant hard instances, we adopt large-scale random masking while utilizing a global recycle network to mitigate the risk of losing key features.
Furthermore, the student updates the teacher using an exponential moving average, which identifies new hard instances for subsequent training iterations and stabilizes optimization. Experimental results on cancer diagnosis, subtyping, survival analysis tasks, and 12 benchmarks demonstrate that MHIM-MIL outperforms the latest methods in both performance and efficiency. The code is available at: \url{https://github.com/DearCaat/MHIM-MIL}.}




\maketitle

\begin{figure*}[t]
\centering
\includegraphics[width=0.95\linewidth]{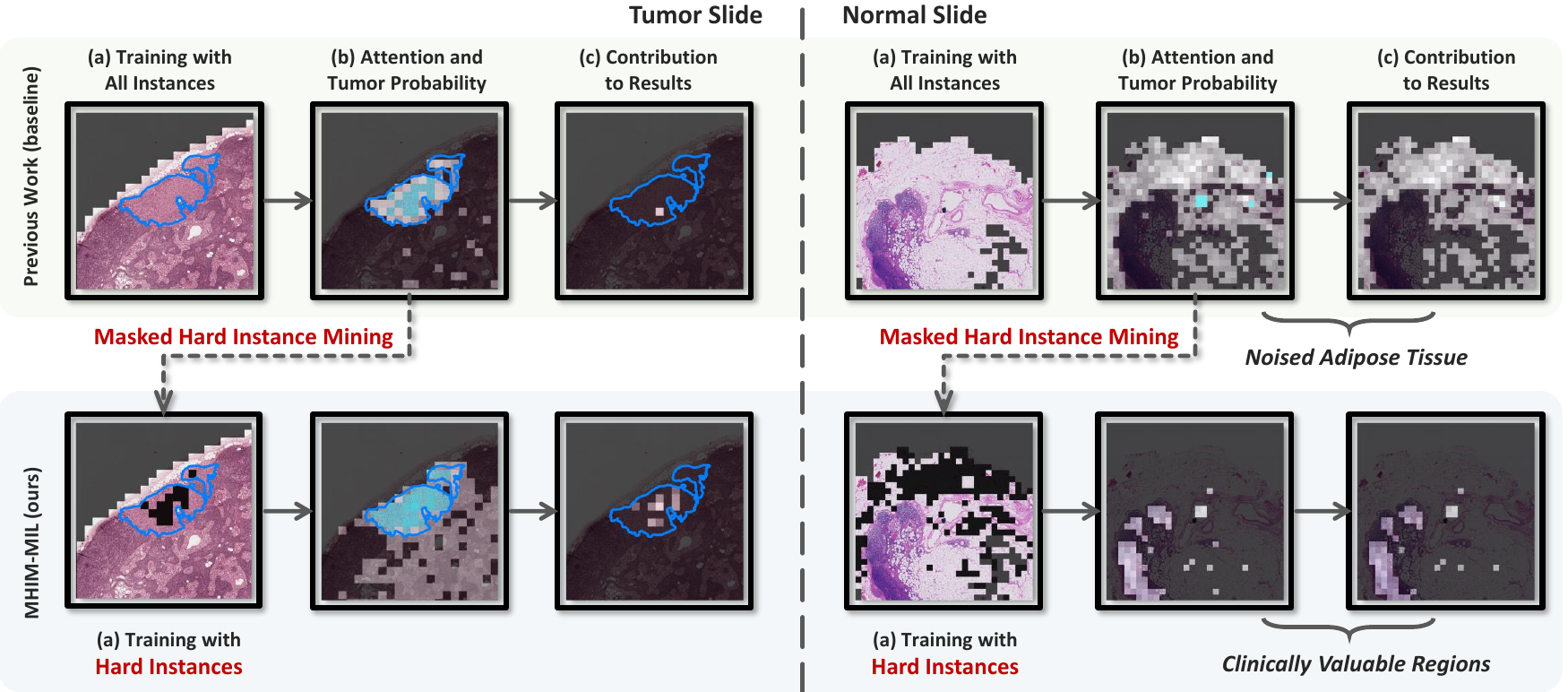}
\caption{
Comparison of MHIM-MIL with baseline methods. The \textbf{\textcolor{blue}{blue}} outlines denote tumor regions annotated by pathologists; subfigure (a) shows masked instances as \textbf{\textcolor{black}{black}} patches. Subfigure (b) displays attention scores from the student model, where brighter patches indicate higher saliency (i.e., mined salient instances), while subfigure (c) presents the softmax-normalized attention weights, with \textbf{\textcolor{cyan}{cyan}} indicating high probability of tumor presence, which should ideally align with the \textcolor{blue}{blue} tumor boundaries. Baseline methods tend to focus on highly activated or trivial regions (top part of subfigure (b) in the tumor slide) or noisy activations (top-right of normal tissue slides); in contrast, MHIM-MIL suppresses such dominant patches via masking, encouraging exploration of less-salient yet informative areas (bottom part of subfigure (b)), leading to more comprehensive and robust predictions.
}
\label{fig:intro_1}
\end{figure*}

\section{Introduction}
\label{sec:introduction}

Computational pathology (CPath), a representative branch in gigapixel image analysis, is an interdisciplinary field that advances computational methods to analyze and model gigapixel histopathology images~\cite{hosseini2024computational}. In modern medicine, histopathology image analysis plays a crucial role, especially in precision oncology, where it is the gold standard for cancer diagnosis~\cite{pinckaers2020intro_pami,li2021dt,zhao2022setmil,lu2021nature}, subtyping~\cite{lin2023interventional}, and prognosis~\cite{yao2020whole,di2022generating_cpath_pami}. Digitizing pathological images into Whole Slide Images (WSIs) has opened new avenues for CPath~\cite{shao2021transmil,chikontwe2021dual}. Due to the huge size of a WSI and the lack of pixel-level annotations, histopathological image analysis is commonly framed as a multiple instance learning (MIL) task~\cite{srinidhi2021survey,dietterich1997mil_2}. In MIL, each WSI (or slide) is considered a bag containing thousands of unlabeled instances (patches) cropped from the slide. If at least one instance is positive, the bag is labeled positive; otherwise, it is labeled negative.


However, the number of slides is limited, and each slide contains a large number of instances with a low positive proportion. This imbalance hinders the inference of bag labels~\cite{zhang2022dtfd,ilse2018attention}. To address this issue, several CPath methods~\cite{ilse2018attention,shao2021transmil,li2021dual,li2021dt,chen2022hipt} employ attention mechanisms to aggregate salient instance features into a bag-level feature for CPath. Figure~\ref{fig:intro_1} demonstrates the superior capability of these methods in mining salient instances from a large number of instances. Furthermore, some MIL frameworks~\cite{zhang2022dtfd,li2021dual,clam,xu2019camel} continue to advance the concept of salient instance mining and focus on how to utilize it to facilitate CPath. For instance, existing frameworks~\cite{zhang2022dtfd,xu2019camel} propose selecting only the instances that correspond to the top $K$ highest or lowest attention scores~\cite{xu2019camel,li2021dual} or patch probabilities~\cite{zhang2022dtfd} to yield high-quality bag embeddings for both training and testing.


These salient instances are essentially "easy-to-classify" instances, which are suboptimal for training a discriminative MIL model. The left part of Figure~\ref{fig:intro_1} demonstrates that previous MIL models tend to focus on salient regions. However, they fail to effectively utilize this information. Instead, they neglect most features and excessively concentrate on a few salient features during feature aggregation, compromising model discriminativeness. Moreover, the right part further illustrates the potential misleading of these "easy-to-classify" instances, causing the models to be influenced by noisy adipose tissue in its discrimination.

In traditional machine learning, such as Support Vector Machines (SVM)~\cite{hearst1998svm}, samples near the category distribution boundary are more challenging to classify but are more valuable for defining the classification boundary. Furthermore, other deep learning studies~\cite{sun2019him_3,tan2022him_4,schroff2015facenet} have shown that incorporating hard samples in training can enhance model generalization. By applying this idea at the instance level, we can better highlight the "hard-to-classify" instances that facilitate MIL model training and benefit the tasks in CPath. However, the absence of instance labels presents a challenge in applying traditional hard sample mining directly at the instance level.

To address this issue, we propose a novel MIL framework based on masked hard instance mining strategies (MHIM), called MHIM-MIL. The core concept of MHIM is to mask easy-to-classify instances, thereby emphasizing hard instances for model training. We introduce a momentum teacher model to evaluate all instances and propose a class-aware instance probability for more accurate assessment of easy-to-classify instances. 
Unlike class-agnostic attention that focus solely on salient features without encoding any class information, this approach directly measures the classification probability of instances.
Consequently, the teacher model effectively mines hard instances for training the MIL model (i.e., the student model). Importantly, the momentum teacher is updated using an exponential moving average (EMA) of the student model. In contrast to conventional MIL frameworks~\cite{zhang2022dtfd,xu2019camel} that employ complex cascade gradient-updating structures, our method is simpler and has less parameters. This approach not only enhances efficiency but also improves performance stability. Furthermore, MHIM is optimized by introducing a consistency constraint that enhances the student's ability to extract discriminative features from hard instances.

The notable feature of MHIM is its ability to provide high-quality hard instances for MIL model training. In scenarios with limited slides and an abundance of patches, the hard instance sequence ideally should be non-redundant, diverse, sufficiently challenging, and error-free. 
To achieve this, we mask out a small portion (e.g., $<$ 5\%) of the easiest-to-classify instances and then introduce a large proportion (e.g., 70\% $\sim$ 90\%) of random masking. This strategy not only significantly reduces sequence length but also enhances the diversity of hard instance sequences, improving both training efficiency and model generalization. Moreover, to mitigate the potential risk of losing key features due to our large-scale random masking strategy, we propose the Global Recycle Network (GRN) to recover these critical features from randomly masked instances at a global level.

We validate MHIM on various CPath tasks, including cancer diagnosis (CAMELYON~\cite{bejnordi2017diagnostic,bandi2018detection}), subtyping (TCGA-NSCLC, TCGA-BRCA), and survival analysis (TCGA-LUAD, TCGA-LUSC, TCGA-BLCA). To demonstrate the framework's generalizability, we employ multiple feature extractors (ResNet-50~\cite{he2016deep}, PLIP~\cite{huang2023visual}, UNI~\cite{chen2024towards}) and baseline models (AB-MIL~\cite{ilse2018attention}, TransMIL~\cite{shao2021transmil}, DSMIL~\cite{li2021dual}). Under multi-fold cross-validation settings, MHIM shows significant improvements across all three baseline models and achieves state-of-the-art performance on multiple benchmarks. For instance, MHIM (TransMIL) improves the C-index by 1.8\% and 1.5\% compared to the baseline and the second-best model, respectively, on TCGA-BLCA-UNI. Furthermore, we shows the advantages of MHIM's computational cost. MHIM (TransMIL) reduces training time and memory consumption by 20\% and 50\%, respectively.

Parts of this paper were published originally in ICCV 2023~\cite{tang2023mhim}. However, we refer to this version as MHIM-v2 and extend our earlier work in several important aspects:

\begin{itemize}
     \item We improve instance assessment by introducing class-aware instance probability, which enables the teacher to more accurately mine hard instances compared to attention scores. 
     \item We simplify the masked hard instance mining strategy to improve framework generalization and incorporate a Global Recycling Network (GRN) to reuse randomly masked features. Extensive experiments show that the GRN boosts the performance of various baseline models without significantly affecting efficiency. 
     \item We extend our MHIM framework to Survival Analysis, providing a more comprehensive evaluation of its impact on CPath. 
     \item We expand the dataset for cancer diagnosis and subtyping tasks, introduce a feature extraction network based on a multimodal pathology large model, and standardize all experiments to 5-fold cross-validation, enabling a more systematic and general validation of the MHIM. 
    
\end{itemize}

\section{Related Work}

In this section, we review the related work from two main perspectives: Computational Pathology and Hard Sample Mining. We first provide a comprehensive overview of computational pathology, detailing the dominant MIL paradigm, its variants, clinical applications, multi-modal extensions, and common instance sampling strategies. We then discuss the concept of hard sample mining in the broader computer vision field, which forms the core inspiration for our approach to address the limitations of existing MIL-based pathology methods.

\subsection{Computational Pathology}
The shift from traditional glass slides to digital pathology has unlocked a vast array of possibilities for computational pathology. This emerging field seeks to integrate the expertise of pathology with advanced image analysis and cutting-edge computer science methodologies to create sophisticated, computer-assisted tools for the interpretation and analysis of pathology images. By leveraging these technologies, computational pathology not only enhances diagnostic accuracy but also facilitates large-scale data analysis, enabling the discovery of novel biomarkers and the development of personalized treatment strategies~\cite{cui2021artificial, song2023artificial,fan2024learning_ijcv_cpath,xiong2025repsnet_ijcv_cpath,zheng2024deep_ijcv_cpath}. Given the gigapixel resolution of WSIs, which makes direct processing infeasible and fine-grained annotation prohibitive, the field has widely adopted a weakly-supervised learning paradigm to effectively analyze these massive datasets.

\noindent\textbf{Multiple Instance Learning in Computational Pathology. }
Among various methodologies, Multiple Instance Learning (MIL)~\cite{dietterich1997mil_2} has emerged as the de facto standard~\cite{li2021dual,song2024morphological,jaume2024transcriptomics,nasiri2024vim4path,li2024virtual}.
MIL is a weakly supervised learning framework that utilizes coarse-grained bag labels for training instead of fine-grained instance annotations.
Previous works can be broadly categorized into two groups: instance-level~\cite{campanella2019clinical,hou2016patch,javed2024unsupervised,instance_mil_1,qu2024rethinking} and embedding-level~\cite{chikontwe2021dual,wu2021combining,li2024dynamic,sharma2021cluster,wang2018revisiting,tang2025revisiting}.
The former obtains instance labels and aggregates them to obtain the bag label, whereas the latter aggregates all instance features into a high-level bag embedding for bag prediction. Due to the lack of instance labels, recent instance-level methods focus on generating pseudo-labels. For example, Qu et al.~\cite{qu2024rethinking} proposed an accurate pseudo-label generator through prototype learning within weakly supervised contrastive learning, which effectively facilitates instance-level learning.
However, these methods have become less prevalent as their performance is fundamentally bottlenecked by the quality of the generated pseudo-labels, which are often noisy and unreliable under weak bag-level supervision.
Conversely, most embedding-level methods~\cite{li2021dual,shao2021transmil,zhang2022dtfd} share the basic idea of AB-MIL~\cite{ilse2018attention}, which employs learnable weights to aggregate salient instance features into a bag embedding.
Furthermore, some MIL frameworks~\cite{zhang2022dtfd,li2021dual,clam,xu2019camel} mine more salient instances, making classification easier and facilitating model learning.
For example, Lu et al.~\cite{clam} selected the most salient instances based on their attention scores to compute instance-level loss and improve performance.
Zhang et al.~\cite{zhang2022dtfd} proposed a CAM-1D based on the AB-MIL paradigm to better mine salient instances and used AB-MIL to aggregate them into a bag embedding.
Recent Mamba-based MIL~\cite{zhang20252dmamba,zheng2025m3amba,yang2024mambamil} frameworks leverage the Mamba state-space model to efficiently capture long-range dependencies in gigapixel WSIs, enabling linear-time integration of global context. However, they still require specialized designs to mitigate the inherent constraints of sequential modeling.
In addition, inspired by CLIP~\cite{clip}, vision-language-based methods~\cite{chen2022hipt,huang2023visual,shi2024vila,li2024generalizable} have recently emerged. These methods explore how to generate and utilize prompts to capture salient instances from thousands of image patches. For example, Shi~et al.~\cite{shi2024vila} proposed a dual-scale visual descriptive text prompt based on a frozen large language model to effectively boost the performance of vision-language models. However, these methods focus excessively on salient instances during training, which are easy instances with high confidence scores and can be easily classified. Consequently, they overlook the importance of hard instances for training. In this paper, we aim to mine hard instances to improve CPath performance.

\noindent\textbf{Clinical Tasks on Computational Pathology.} Building upon the MIL framework, CPath has achieved significant progress in various clinical applications. Currently, computational pathology is widely applied in various tasks including automated tumor diagnosis, subtyping, survival analysis, grading of cancers, and quantification of histopathological features, thereby significantly enhancing diagnostic consistency and reducing the workload on pathologists.
Among these tasks, classification and survival prediction stand out as two of the most crucial applications.
Classification, which encompasses diagnosis and subtyping, focuses on accurately identifying and categorizing different types of cancers based on histopathological images~\cite{ilse2018attention,clam,li2021dual,shao2021transmil,zhang2022dtfd,lin2023interventional,tang2023mhim}. This is essential for determining the appropriate course of treatment and has been significantly advanced by deep learning techniques, which can discern subtle patterns and features in pathology slides that may not be easily detected by the human eye.
Survival analysis, a critical task in cancer prognosis, aims to assess the probability of an event (typically death) occurring for a particular patient and accurately rank the risks of cancer patients. It is indispensable in computational pathology, providing insights into disease progression, treatment effectiveness, and patient prognosis. This task can be expressed as an estimation of the hazard function, and early classical methods~\cite{dickson1989prognosis,ohno1997comparison} were based on Cox's regression model~\cite{cox1972regression}. Recently, deep learning-based methods have become mainstream, with some recent works~\cite{zhu2017wsisa,yao2020whole,di2022generating_cpath_pami} constructing effective survival analysis models based on pathological images. These methods also rely on the MIL paradigm, attempting to utilize AB-MIL to focus on salient regions in WSI. For example, Yao et al.~\cite{yao2020whole} introduced the siamese MI-FCN and attention-based MIL pooling to efficiently learn imaging features from the WSI and then aggregate WSI-level information to the patient level.

\noindent\textbf{Multi-modal in Computational Pathology. }
While unimodal approaches using only WSIs have formed the foundation of CPath, a recent and powerful trend is the integration of multiple data modalities to further boost performance and robustness. Integrating genomics and WSIs in multimodal approaches~\cite{zhou2023cross,shao2023fam3l,chen2020pathomic,nakhli2023sparse} has recently attracted increasing attention due to their advantages in performance and robustness. The state-of-the-art method CMTA~\cite{zhou2023cross} has demonstrated that leveraging genomics information and tailored design can outperform unimodal methods by a large margin.
The integration of multiple modalities is not limited to genomics. For example, Song et al.~\cite{song2025deep} proposed a Swin-Transformer fusion model that integrates WSIs with CT images, achieving notably higher prognostic accuracy than either modality alone. Nakhli et al.~\cite{nakhli2023sparse} introduced a graph-based transformer that combines multiple histopathology image modalities per patient, which significantly outperformed prior methods and remained robust even when some modality data were missing. Other approaches incorporate clinical attributes or textual pathology reports alongside WSIs to inject domain knowledge and improve interpretability~\cite{xu2024multimodal}, highlighting the promise of vision–language and knowledge-integrated models in this field. These multimodal studies underscore the potential of fusing heterogeneous data sources for more robust and informative pathology AI models.
However, despite the significant performance advantages of multimodal methods, obtaining high-quality multimodal data is more challenging. We aim to enhance the performance of unimodal methods in computational pathology to narrow the gap with multimodal methods, providing a lower-cost and more generalizable option.

\begin{figure*}[t]
\centerline{\includegraphics[width=0.8\textwidth]{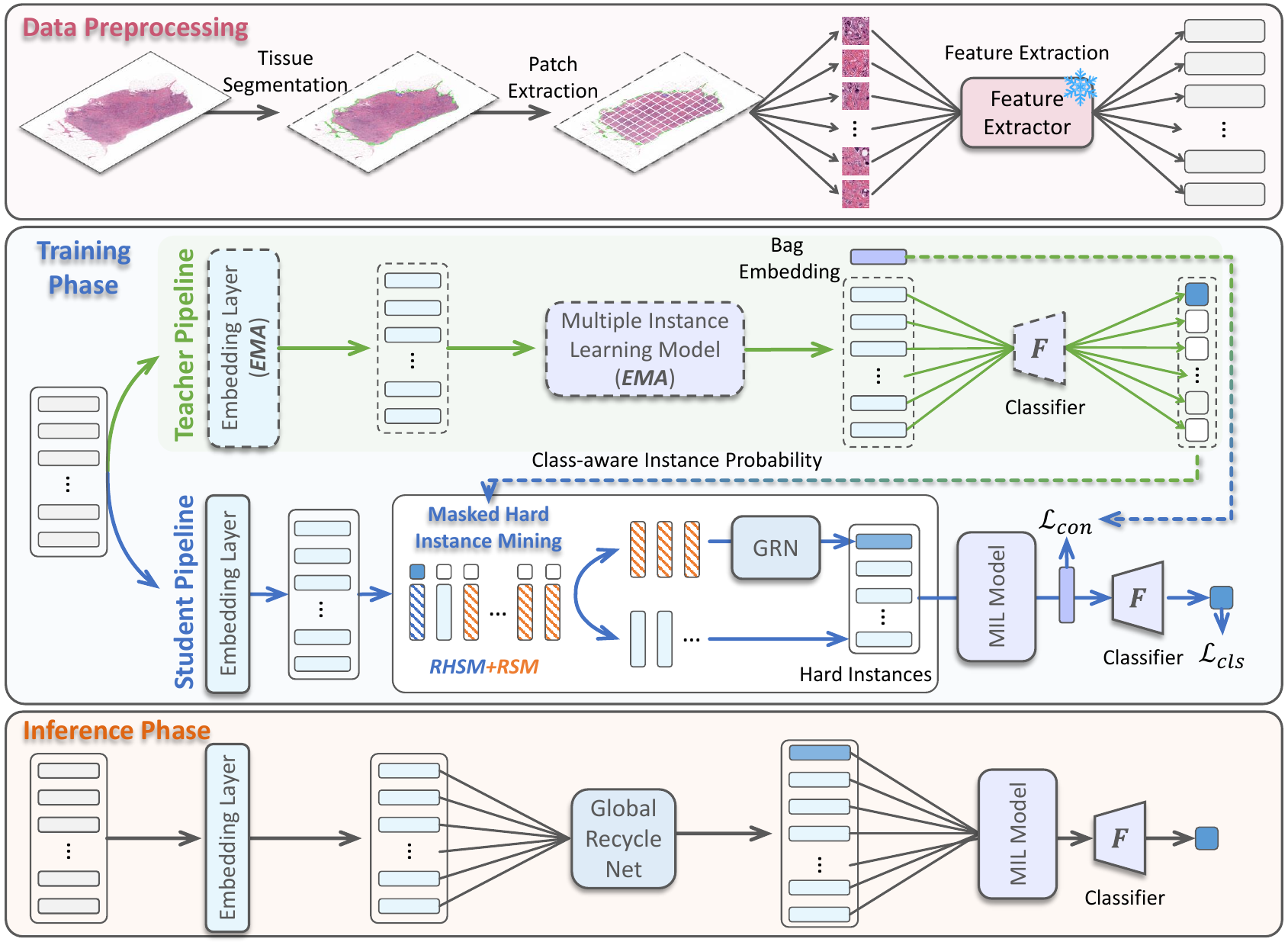}}
\caption{Overview of proposed MHIM-MIL. A momentum teacher is used to compute class-aware instance probability for all instances. We mask instances based on these probabilities using randomly high score masking (RHSM) and random score masking (RSM) strategies. Then, we employ a global recycle network (GRN) to reconstruct key features from the large-scale RSM-masked instances. Finally, we feed these reconstructed features along with unmasked instances to the student model. The student is updated using a consistency loss term $\mathcal{L}_{con}$ and a label error loss term $\mathcal{L}_{cls}$. The teacher parameters are updated with an Exponential Moving Average (EMA) of the student parameters without gradient updates. During inference, we use the complete input instances and the student model only.
}
\label{fig:model}
\end{figure*}

\noindent\textbf{Instance Sampling in Computational Pathology.}
MIL-based computational pathology aims to identify the most informative features and predict the final result. However, it is non-trivial to mine high-quality features from thousands of instances, and long sequence inputs pose significant memory challenges for Transformer models. Consequently, many methods attempt to sample long instance sequences to reduce problem complexity. Random sampling is a classic method, and~\cite{lin2023interventional,shao2023lnpl,zhu2022murcl} use random strategies to divide the total instances into several groups, referred to as pseudo-bags. Then, the bag labels are assigned to the pseudo-bags as pseudo-labels. Although this simple and effective strategy can achieve substantial improvement, it also faces serious noise label issues. To alleviate these issues, some efforts~\cite{chen2023rankmix,yang2023protodiv} employ attention mechanisms to improve sampling and pseudo-bag labeling. However, the pseudo-bag method still cannot effectively solve the efficiency dilemma caused by long sequence inputs. Therefore, recent works~\cite{zhang2023multi,li2021dt,zhao2022setmil} directly sample the instance sequence randomly and discard the remaining parts to enable Transformer model training. In addition, feature clustering methods~\cite{Zhang_2022_BMVC,sharma2021cluster,zhou2024end} compute cluster centroids of all feature embeddings and sample representative embeddings for the final prediction. Moreover,~\cite{zheng2024dynamic} utilizes reinforcement learning to propose an iterative dynamic instance sampling strategy to improve instance selection and feature aggregation. While our method shares similar spirit with the above approaches, we focus on how to simply and efficiently mine high-quality hard instances to facilitate model training and significantly enhance training efficiency.

\subsection{Hard Sample Mining}
Hard sample mining is a popular technique to speed up convergence and enhance the discriminative power of the model in many deep learning areas, such as face recognition~\cite{schroff2015facenet,lee2024hard}, object detection~\cite{shrivastava2016training,wang2018hsm}, person re-identification~\cite{sun2019him_3,tan2022him_4,rao2024hierarchical_ijcv_him,rao2024hierarchical}, graph representation learning~\cite{perozzi2014deepwalk,Tang_Qu_Wang_Zhang_Yan_Mei_2015,huang2021mixgcf,niu2024affinity}, deep metric learning~\cite{suh2019him_2,yan2024causality}, and more~\cite{wang2024not,kheradmand2024accelerating,sahin2024enhancing}. 
The main idea behind this technique is to select the samples which are hard to classify correctly (i.e., hard negatives and hard positives) for alleviating the imbalance between positive and negative samples and facilitating model training\cite{yang2024does}. 
There are generally three groups of approaches for evaluating sample difficulty: loss-based~\cite{hermans2017defense}, similarity-based~\cite{chen2017beyond}, and learnable weight-based~\cite{xu2019learning}. 
Typically, these strategies require complete sample supervision information. 
Drawing on the ideas of the above works, we propose a hard instance mining approach in MIL, mining hard examples at the instance level. 
In this, there are no complete instance labels, only the bag label is available.
Similar to our approach, Li~et al.~\cite{li2019hnm} utilized attention scores to identify salient instances from false negative bags to serve as hard negative instances and used them to compose the hard bags for improving classification performance. 
A key difference is that we indirectly mine hard instances by masking out the most salient instances rather than directly locating hard negative instances.


\section{Proposed Method}

\subsection{Background: MIL Formulation}
In MIL, a WSI $X$ is modeled as a bag of instances, represented as $X = \left\{x_i\right\}^N_{i=1}$, where $x_i$ is a patch extracted from the WSI and $N$ is the total number of instances. For a classification task, a known bag-level label $Y \in C$ is available, while the instance-level labels $y_i \in C$ are unknown. The objective of a MIL model $\mathcal{M}(\cdot)$ is to predict the bag label based on all instances: $\hat{Y}\leftarrow\mathcal{M}(X)$.
While classical instance-level methods derive the bag label from instance-level predictions, our focus is on a more prevalent embedding-level approach that involves an instance aggregation step. This process creates a bag representation $F$ from the extracted features of all instances in the bag, $Z=\{z_i\}_{i=1}^N$. A classifier $\mathcal{C}(\cdot)$, trained on this representation, is then used to predict the bag label, $\hat{Y}\leftarrow\mathcal{C}(F)$. There are two primary strategies for instance aggregation, one of which is the attention-based aggregation method~\cite{ilse2018attention}, denoted as follows,
\begin{equation}
F = \sum_{i=1}^{N}a_{i}z_{i} \in \mathbb{R}^{D},
\end{equation}
where $a_{i}$ is the learnable scalar weight for $z_i$, and $D$ is the dimension of vector $F$ and $z_i$. Many embedding-level works~\cite{li2021dual,clam,zhang2022dtfd} follow this formulation but differ in the ways they generate the attention score $a_i$. Notably, DSMIL~\cite{li2021dual} predicts instance labels during the process of aggregating bag embedding, with attention scores relying on the predicted instance labels.

Another is the multi-head self-attention (MSA) based aggregation~\cite{shao2021transmil}. In this approach, a class token $z_{0}^{0}$ is combined with the instance features to form the initial input sequence $ Z^0=\left[z_{0}^{0},z_1^{0},\dots,z_N^{0}\right]\in \mathbb{R}^{\left(N+1\right) \times D}$ for aggregating instance features. With the single attention head $\text{head}_h$, 
this can be formulated as,
\begin{equation}
\begin{aligned}
&\text{head}_h = A^{\ell}_{h} \left (Z^{\ell-1}W^{V}_{h}\right ) \in \mathbb{R}^{(N+1)\times \frac{D}{H}}, &\ell=1\dots L,\\
&Z^{\ell} =  \textrm{Concat}\left ( \textrm{head}_{1},\cdots, \textrm{head}_{H}\right )W^O, &\ell=1\dots L,
\end{aligned}
\end{equation}
where $W^{V}\in \mathbb{R}^{D \times \frac{D}{H}}$ and $W^{O}\in \mathbb{R}^{D \times D}$ are the learnable projection matrices of MSA. $A^{\ell}_h\in \mathbb{R}^{(N+1)\times (N+1)}$ is the $h$-th head attention matrix of the $\ell$-th layer, $L$ is the number of MSA block, and $H$ is the number of head in each MSA block. The bag embedding $F$ is the class token of the final layer,
\begin{equation}
F = z_{0}^{L}.
\end{equation}
It is to be noted that the self-attention-based aggregation fundamentally represents a variant of attention-based MIL aggregation. This approach excels in multiple computational pathology tasks due to its superior long-distance modeling capabilities. However, it also faces challenges related to efficiency due to the processing of long-sequence inputs.
Collectively, these methods can be referred to as the \textbf{\textit{general attention-based MIL.}}
\setlength{\algomargin}{0em}
\SetAlFnt{\small}
\SetAlCapNameFnt{\small}

\begin{algorithm}[t]
  \caption{Hard Instance Mining}
  \label{algo:mhim}
  \setstretch{0.8}
  \PyComment{m\underline{~}t: the MIL model of teacher networks} \\
  \PyComment{c\underline{~}t: the classifier of teacher network} \\
  \PyComment{mhr: high score mask ratio} \\
  \PyComment{mrr: random mask ratio} \\
   \PyComment{q\underline{~}g: global query} \\
  \PyComment{mm\underline{~}q: momentum rates of global query} \\
  ~\\
  \PyCode{def mhim\underline{~}fn(x):}\\
  \Indp
  \PyComment{easy instance assessment} \\
  \PyCode{attn,f\underline{~}ins = m\underline{~}t(x)}\\
  \PyCode{f\underline{~}ins = einsum("nd,n->nd",f\underline{~}ins,attn)}\\
  \PyCode{score = c\underline{~}t(f\underline{~}ins)}\\
   ~\\
  \PyComment{high score masking} \\
  \PyCode{score = sort(score)}\\
  \PyCode{ids\underline{~}h = topk(score,mhr*score.length)}\\
  \PyCode{z,\underline{~} = mask(x,ids\underline{~}h)}\\
  \PyComment{high score mask ratio decay}\\
  \PyCode{cosine\underline{~}decay(mhr)}\\
  \PyComment{large scale masking} \\
  \PyCode{ids\underline{~}r = random\underline{~}select(z,mrr*z.length)}\\
  \PyCode{z,z\underline{~}m = mask(z,ids\underline{~}r)}\\
  ~\\
  \PyComment{global recycle network} \\
  \PyCode{g\underline{~}q.require\underline{~}gradient = False}\\
  \PyCode{z\underline{~}m = mca(g\underline{~}q,z\underline{~}m)}\\
  \PyComment{update global query via ema} \\
  \PyCode{g\underline{~}q = mm\underline{~}q*g\underline{~}q+(1-mm\underline{~}q)*z\underline{~}m)}\\
  ~\\
  \PyComment{obtain the final hard instances for student model} \\
  \PyCode{z = concat(z,z\underline{~}m, dim=0)}\\
  \PyCode{return z}\\
  \Indm
\end{algorithm}

\subsection{MHIM-MIL for CPath}
In general attention-based MIL frameworks, the attention score typically serves as a criterion to evaluate the importance of instances, denoting their contributions to bag embedding.
The salient instances with high scores are useful for classifying WSI in the testing phase, but are not conducive to training a MIL model with good generalization ability. 
Although hard samples have been proven to enhance the generalization ability of the model in many computer vision scenarios~\cite{dong2017him_1,tan2022him_4,suh2019him_2,sun2019him_3},
previous MIL work focuses more on exploiting salient instances and neglecting the utilization of hard instances in model optimization.

In this paper, we propose a simple and efficient MIL framework with Masked Hard Instance Mining (MHIM-MIL) to enhance CPath.
As illustrated in Figure~\ref{fig:model}, the MHIM-MIL framework incorporates a Siamese structure during training. The backbone of our framework is a general attention-based MIL model (referred to as Student), denoted as $\mathcal{S(\cdot)}$, which aggregates instance features. To increase the difficulty of training and encourage the student model to focus on hard instances, we introduce a momentum teacher, represented as $\mathcal{T(\cdot)}$. The teacher assesses easy-to-classify instances using the proposed class-aware instance probability. Class-aware instance probability provides a better evaluation of easy instances and facilitates hard instance mining compared to class-agnostic attention scores. Using these assessment results, the teacher obtains high-quality sequences of hard instances through a masked hard mining strategy. Specifically, it first applies small-scale high-score masking followed by large-scale random masking to reduce redundancy and enhance sequence diversity. To mitigate the risk of losing critical features due to large-scale random instance masking, the student model incorporates a recycle network $\mathcal{R}(\cdot)$ to recover crucial features.
Finally, the student model predicts bag labels by concatenating mined hard features and recycled features. 

The teacher model shares the same network structure as the student model, except for the additional recycle network, and does not require gradient-based updates. Due to the varying number of instances within each bag, a non-batch gradient descent algorithm (i.e., SGD with batch size 1) is used to optimize the MIL model. Consequently, unlike traditional MIL frameworks that employ double-tier gradient updating models~\cite{zhang2022dtfd,xu2019camel}, this Siamese structure enables more stable and efficient training with fewer parameters.
The proposed framework is formulated as: 
\begin{equation}
\hat{Y} = \mathcal{S}\left ( \widetilde{Z} \right ) = \mathcal{S}\left ( \mathcal{R}\left (\mathcal{M}_{\mathcal{T}} \left ( Z \right ) \right ) \right ),
\end{equation}
where $\mathcal{M}_{\mathcal{T}}(\cdot)$ denotes the masked hard instance mining executed through the teacher model, and $\widetilde{Z}$ represents the final processed instances.


\begin{figure}[t]
\centering
\includegraphics[width=\linewidth]{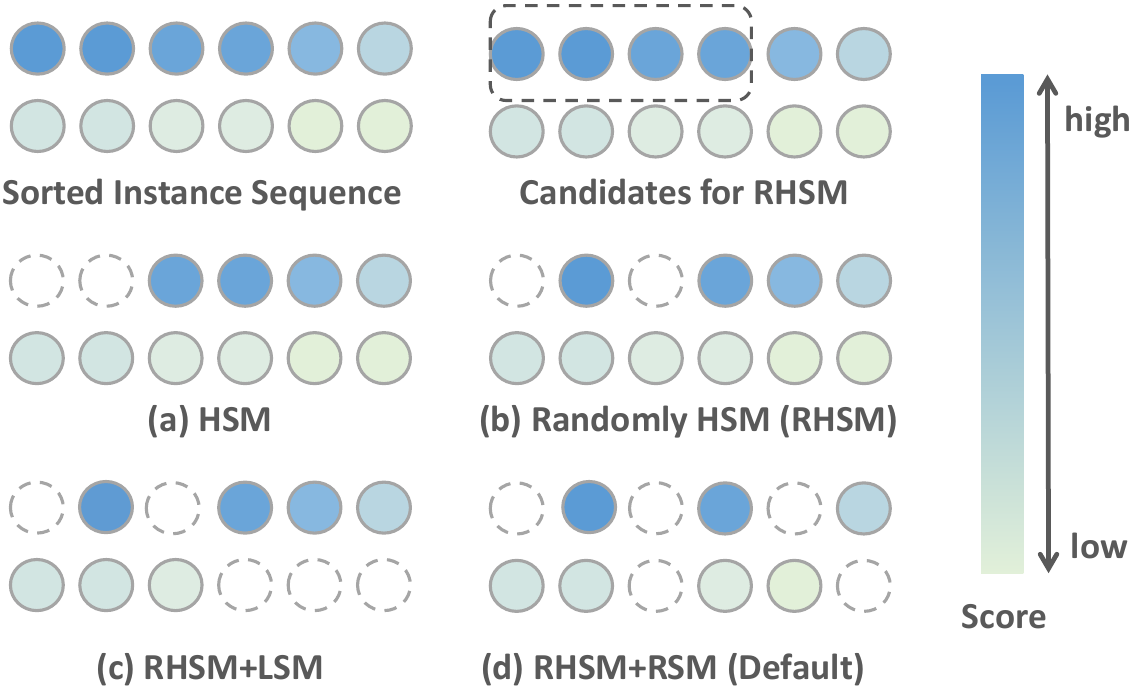}
\caption{Illustration of various hard instance mining methods.}
\label{fig:hsm}
\vspace{-0.3cm}
\end{figure}

\subsection{Masked Hard Instance Mining}
Traditional hard sample mining strategies often struggle in the absence of instance-level supervision. To tackle this challenge, we propose a masked hard instance mining approach that leverages assessment scores to implicitly identify difficult instances by masking those that are easier with higher assessment scores.

\subsubsection{Easy Instance Assessment}
The teacher assesses each instance to identify the easy ones. More specifically, given a complete sequence of instance features $Z =  \{ z_i\}^{N}_{i=1}$ as the input of the teacher $\mathcal{T}\left ( \cdot  \right )$, the teacher outputs the attention score $a_i$ for each instance as follow,
\begin{equation}
    A = \left [ a_1,\dots,a_i,\dots,a_N \right ] = \mathcal{T}\left ( Z  \right ).
\end{equation}
The attention score indicates focus level of the MIL model on each instance, making it an intuitive and commonly used metric for evaluation. However, it does not explicitly show the contribution of each instance to the classification outcome. As highlighted in literature~\cite{zhang2022dtfd}, instances with higher attention scores may contribute insignificantly to the diagnosis. Inspired by this observation, we introduce the Class-aware Instance Probability to evaluate instance more accurately within a general attention-based MIL model. 

Specifically, for the teacher that include an instance classifier, such as DSMIL~\cite{li2021dual}, we utilize the instance classifier to classify the weighted features derived from multiplying attention weights by instance features. For teachers without an instance classifier, we employ the bag classifier for this purpose. This process is defined as,
\begin{equation}
    S = \mathcal{C}_{\mathcal{T}} \left ( A \cdot Z  \right ),
\end{equation}
where $\mathcal{C}_{\mathcal{T}}(\cdot)$ is the instance classifier of the teacher model. $S$ is the obtained class-aware instance probabilities and $A$ is the attention weights. This computation does not introduce additional networks and effectively enhances the accuracy of the general attention-based MIL model in easy-classify instances assessment. Further discussions on this can be found in the experimental section.

Then, we obtain the indices of the score sequence in descending order by applying a sorting operation on $S$,
\begin{equation}
    I = \left [ i_{1},i_{2},\dots,i_{N} \right ] = \textrm{Sort}\left ( S  \right ),
\end{equation}
where $i_{1}$ is the index of the instance with the highest score, while $i_N$ is the index of the one with the lowest score. With this index collection $I$, we introduce several strategies for masked hard instance mining to identify hard instances. An $N$-dimensional binary vector $M=[m_1,\dots,m_i,\dots,m_N]$ is defined to encode the masking flag of instances, where $m_i\in\{0,1\}$. An instance is masked if $m_i=1$, otherwise, it is considered unmasked.

\subsubsection{Randomly High Score Masking}
The simplest yet most critical strategy in masked hard instance mining is the High Score Masking (HSM) strategy, which masks instances with the top $\beta_h$\% highest assessment scores. For HSM, the instance mask flags are initially set to zero vectors, $M_{h}(:)=0$. We then select the indices of instances with scores in the top $\beta_h$\%, denoted by $I_h=[i_{t}]_{t=1}^{\left \lceil \beta_h\% \times N \right \rceil}$. These indices are used to update the mask flags, setting $M_h(I_h)=1$.
To ensure that positive instances are preserved within the unmasked sequences, we also utilized techniques such as mask ratio decay. 
The ratio $\beta_h$\% is nonlinearly decreased using a cosine function as training progresses. Despite these measures, HSM may face a major challenge: it could mask all distinctive features in the early stages of training, leading to ``error instance mining''. To mitigate this, we introduce the Randomly High Score Masking technique, which selects instances with the top $2\times \beta_{h}$\% attention scores as potential candidates and randomly masks half of them to preserve distinctive information. Figure~\ref{fig:hsm} illustrates this approach.
The mined instance sequence can be obtained by applying the high score masking operation,
\begin{equation}
    \hat{Z}_{h} \leftarrow \mathcal{M}_{\mathcal{T}}^{h} \left ( Z \right ) = \textrm{Mask}\left ( Z, M_{h} \right ) \in \mathbb{R}^{\hat{N}\times D},
\end{equation}
where the $\hat{N} = N - \left \lceil \beta_h\% \times N \right \rceil$ is the number of unmasked instances.


\begin{figure}[tb]
\centering
    \includegraphics[width=0.97\linewidth]{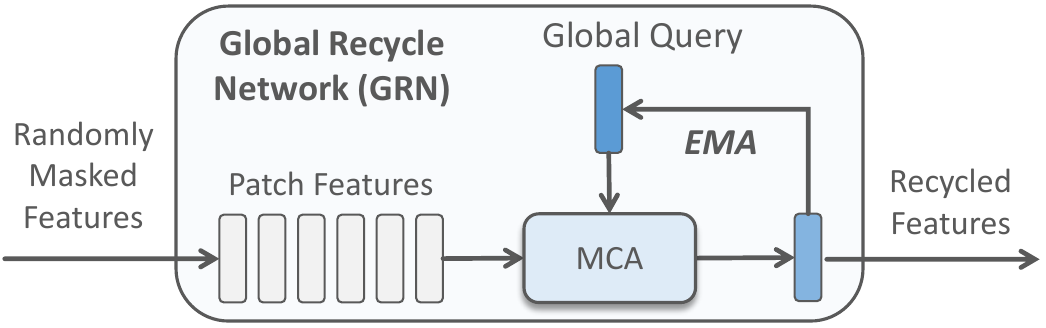}
    \caption{Illustration of Global Recycle Network.}
    \label{fig:recycle}
\end{figure}

\subsubsection{Large Scale Masking}

Although HSM demonstrates effectiveness in mining hard instances, it is limited by a low masking ratio (e.g., $< 5\%$), which fails to significantly improve the efficiency of the MIL model. Over-length sequences not only interfere with the training of student models but also hinder the application of advanced aggregation mechanisms, such as Multi-Head Self-Attention (MSA). To address this, we apply high ratio (e.g., $70 \% \sim 90\% $) masking to the instances mined by HSM, as depicted in Figure~\ref{fig:hsm} (c) and (d). We propose the following two strategies to refine hard instance mining and achieve more properties:
\begin{itemize}
    \item \textbf{Low Score Masking (LSM)}: 
     We use the same pipeline as HSM to generate the mask flags $M_{l}$ for masking the instances with the top $\beta_l$\% lowest attention scores in order to filter out the redundant 
     uninformative instances and improve efficiency. 
    \item \textbf{Random Score Masking (RSM)}: Randomness is beneficial to reduce the risk of over-fitting. We generate a random mask flag vector $M_{l}$ with a given random ratio $\beta_l$\%, and combine it with HSM for introducing the randomness to the hard instance mining.
\end{itemize}

Once the large ratio masking flag $M_{l}$ is produced, we can obtained new mined instance sequence $\hat{Z}_{l} \in \mathbb{R}^{\left \lfloor \left ( 1-\beta_{l} \right ) \times \hat{N} \right \rfloor\times D}$ and masked sequence $\hat{Z}_{m}\in \mathbb{R}^{\left \lceil \beta_{l} \times \hat{N} \right \rceil\times D}$,
\begin{equation}
    \{\hat{Z}_{l}, \hat{Z}_{m}\} \leftarrow \mathcal{M}_{\mathcal{T}}^{l} \left ( \hat{Z}_{h} \right ) = \textrm{Mask}\left ( \hat{Z}_{h},M_{l} \right ),
\end{equation}
where $\hat{Z}_{l}\cup\hat{Z}_{m}=\hat{Z}_{h}$.

We believe RSM is the better choice as it can more effectively alleviate over-fitting by increasing image diversity. Moreover, given the inherent sparsity in attention-based MIL models, LSM struggles to further reduce redundancy.


\begin{algorithm}[t]
  \setstretch{0.8}
  \PyComment{f\underline{~}t, f\underline{~}s: teacher and student networks} \\
  \PyComment{f\underline{~}p: the pretrained network} \\
  \PyComment{m: momentum rates} \\
  \PyComment{tp: temperatures} \\
  \PyComment{a: consistency loss scaling factor} \\
  ~\\
  \PyComment{initialize} \\
  \PyCode{f\underline{~}t.params = f\underline{~}p.params}\\
  \PyCode{f\underline{~}s.head.params = f\underline{~}p.head.params}\\
  ~\\
  \PyComment{load a slide and label x,y}\\
  \PyCode{for x,y in loader:} \\
  \Indp   
    \PyComment{get bag embedding from teacher}\\
    \PyCode{\underline{~},be\underline{~}t = f\underline{~}t.forward(x)}\\
    \PyComment{stop gradient of teacher network}\\
    \PyCode{be\underline{~}t = be\underline{~}t.detach()}\\
    ~\\
    \PyComment{masked hard instance mining}\\
    \PyComment{algorithm 1 gives more details}\\
    \PyCode{x\underline{~}hard = mhim\underline{~}fn(x)}\\
    \PyCode{logits\underline{~}s,be\underline{~}s = f\underline{~}s.forward(x\underline{~}hard)}\\
    ~\\
    \PyComment{consistency loss}\\
    \PyCode{loss\underline{~}con = -softmax(be\underline{~}t / tp) * log\underline{~}softmax(be\underline{~}s)}\\
    \PyComment{label prediction loss}\\
    \PyCode{loss\underline{~}cls = cross\underline{~}entropy(logits\underline{~}s,y)}\\
    \PyCode{loss\underline{~}all = loss\underline{~}cls + a*loss\underline{~}con}\\
    ~\\
    \PyComment{gradient descent: student network}\\
    \PyCode{loss\underline{~}all.backward()}\\
    \PyCode{update(f\underline{~}s.params)}\\
    \PyComment{ema update: teacher network}\\
    \PyCode{f\underline{~}t.params = m*f\underline{~}t.params+(1-m)*f\underline{~}s.params}\\
  \Indm 
  \caption{Framework Optimization}
  \label{algo:train} 
\vspace{-0.3cm}
\end{algorithm}

\subsection{Global Recycle Network}
Masking instances at a high ratio inevitably lead to a risk of losing critical features, which in turn impairs the performance of the MIL model. This issue becomes more pronounced when key features are sparse. To address this, we propose a Global Recycle Network (GRN) to recover masked discriminative features from a global perspective. We initialize global queries $Q_G \in \mathbb{R}^{K\times D}$. The Multi-head Cross-Attention (MCA) computation between the masked instance sequence $\hat{Z}_{m}$ and $Q_G$, represented as $\textrm{MCA}\left ( \cdot \right )$, is employed to mine key features. The process is formalized as:
\begin{equation}
\widetilde{Z}_{m} = \textrm{MCA}\left ( Q_G , \hat{Z}_{m} \right ), \widetilde{Z}_{m} \in \mathbb{R}^{K \times D},
\end{equation}
where $K$ is the number of queries. Importantly, $Q_G$ is updated not through gradient descent but via Exponential Moving Average (EMA), with the update formula being $Q_G \leftarrow \lambda_q Q_G +(1-\lambda_q) \widetilde{Z}_{m}$. This approach ensures that the query vector effectively captures key features from a global perspective while maintaining the stability of the global query vector. In such a manner, the final hard instances can be denoted as,
\begin{equation}
    \widetilde{Z}= \mathcal{R}\left ( \hat{Z} \right ) = \textrm{Concat} \left ( \hat{Z}_{l}, \widetilde{Z}_{m} \right ).
\end{equation}

\subsection{Consistent Iterative Learning}
Within the Siamese architecture, the teacher model not only guides the training of the student model but also updates itself with the new knowledge acquired by the student model. This iterative optimization process gradually enhances the mining capabilities of the teacher model and the discriminative power of the student model.
To further facilitate this optimization and help the student model learn more effectively from hard instances, we introduce a consistency loss denoted as $\mathcal{L}_{con}$. This aims to align the bag embedding of the teacher model, which takes the entire instance sequence as input, with that of the student model, which uses only the hard instances. This not only accelerates the convergence of the student model during the initial phase of training but also enhances its ability to extract discriminative features from hard instances. 
The entire optimization process can be further divided into two parts based on the two branches.  

\noindent\textbf{Student Optimization:}
 There are two losses in student optimization. One is the cross-entropy for measuring the bag label prediction loss,
\begin{equation}
   \mathcal{L}_{cls} = Y \textrm{log}\hat{Y} + \left ( 1-Y \right )\textrm{log}\left ( 1-\hat{Y} \right ).
\end{equation}
The other is the representation consistency loss between the bag embedding of student $F_s$ and the one of momentum teacher $F_t$,
\begin{equation}
   \mathcal{L}_{con} = - \textrm{softmax}\left ( F_t / \tau  \right ) \, \textrm{log}\, F_s,
\end{equation}
where the $\tau > 0$ is a temperature parameter.
Overall, the final optimization loss is as follows:
\begin{equation}
    \begin{aligned}
   \{\hat{\theta_s}\}\leftarrow \arg\underset{\theta_s}\min ~\mathcal{L} = \mathcal{L}_{cls}+ \alpha \mathcal{L}_{con},
    \end{aligned}
\end{equation}
where  $\theta_s$ is the parameters of the student network $\mathcal{S}(\cdot)$, and $\alpha>0$ is a scaling factor.

\noindent\textbf{Teacher Optimization:}
The parameters of momentum teacher $\theta_t$ are updated by the EMA of the student parameters.
The update rule is $\theta_t \leftarrow \lambda \theta_t+(1-\lambda) \theta_s $, where $\theta_t$ is the parameters of the teacher network $\mathcal{T}(\cdot)$ and $\lambda$ is a hyperparameter. 
Moreover, the updated teacher model will be used in the next iteration of hard instance mining.

\subsection{Inference}
In the testing stage, only the student model is used for predicting the bag label. As shown in Figure~\ref{fig:model}, the hard instance mining step will not be conducted in the inference phase. In other words, $\hat{Y} = \mathcal{S}\left ( \widetilde{Z} \right )$ where $\widetilde{Z} =\textrm{Concat} \left ( Z, \mathcal{R}(Z) \right ) $.


\begin{table}[t]
	\centering
    \footnotesize
	\caption{Details of the Primary WSI Datasets Used. \# denotes the number of it. APPW is the Average Patch per WSI.}   
	\begin{tabularx}{\linewidth}{p{2.5cm}M{1.1cm}M{1cm}C}
		\toprule
		Name  & \#Case  &\#WSI & \#APPW       \\ \midrule
		CAMELYON   & 370    & 899     &   8108     \\
		TCGA-NSCLC & 956    & 1053    &   10302   \\
		TCGA-BRCA  & 1062   & 1131    &   8746   \\
		TCGA-LUAD  & 478    &  541    &   10629   \\
		TCGA-LUSC  & 478    &  512    &   10395  \\
		TCGA-BLCA  & 376    &  457    &   14364  \\
		\bottomrule
	\end{tabularx}
	\label{tab:wsi_set}
\vspace{-0.3cm}
\end{table}

\begin{table*}[t]
\centering
\caption{\justifying Cancer diagnosis and subtyping results on CAMELYON and TCGA-NSCLC. The highest performance is in bold, and the second-best performance is underlined. 
The Accuracy and F1-score are determined by the optimal threshold.
MHIM(baseline) is the conference version of this paper, and we refer to the improved framework as MHIM-v2(baseline). }
\label{tab:main_comp}
\footnotesize
\begin{tabularx}{\textwidth}{Xp{3.1cm}ccccccc}
\toprule
& {\multirow{2}{*}{Methods}}        
& \multicolumn{3}{c}{CAMELYON}   
&& \multicolumn{3}{c}{TCGA-NSCLC}  \\ 
\cmidrule(lr){3-5} \cmidrule(lr){7-9} 
&& Accuracy                 & AUC                   & F1-score  
&& Accuracy                 & AUC                   & F1-score   \\ 
\midrule

\multirow{13}{*}{\rotatebox{90}{\makecell[c]{ResNet-50}}}
&AB-MIL~\cite{ilse2018attention}  
& 88.09$\pm$1.69            & 91.59$\pm$2.27        & 83.35$\pm$2.74
&& 90.99$\pm$1.93     & 95.32$\pm$1.39     & 90.95$\pm$1.93\\

&CLAM~\cite{clam}        
& 88.20$\pm$2.14            & 91.43$\pm$2.09        & 83.23$\pm$3.07
&& 90.52$\pm$2.08     & 95.37$\pm$1.08     & 90.08$\pm$1.97\\

&DSMIL~\cite{li2021dual}  
& 87.53$\pm$2.24            & 90.41$\pm$2.25        & 82.89$\pm$2.90         
&& 90.23$\pm$1.63     & 95.57$\pm$0.81     & 90.20$\pm$1.63\\

&TransMIL~\cite{shao2021transmil} 
& 88.42$\pm$1.45            & 91.23$\pm$1.48        & 84.05$\pm$1.90
&& 90.04$\pm$1.86     & 94.97$\pm$1.11     & 89.94$\pm$1.73\\

&DTFD-MIL~\cite{zhang2022dtfd}  
& 87.64$\pm$1.76            & 91.49$\pm$1.57        & 82.51$\pm$2.02
&& 89.85$\pm$1.53     & 95.55$\pm$1.47     & 89.60$\pm$1.67\\

&IBMIL~\cite{lin2023interventional}    
& 88.08$\pm$1.68            & 91.50$\pm$2.06        & 83.96$\pm$2.19
&& 90.04$\pm$1.48     & 95.57$\pm$1.13     & 89.73$\pm$1.64\\

&MHIM(AB-MIL)~\cite{tang2023mhim}    
& 89.20$\pm$2.11            & 92.30$\pm$1.79        & 84.89$\pm$3.02
&& 91.27$\pm$2.35     & 96.02$\pm$1.35     & 90.85$\pm$2.53\\

&MHIM(TransMIL)~\cite{tang2023mhim}    
& \underline{89.98$\pm$1.16}            & \underline{93.02$\pm$1.12}        & \underline{86.17$\pm$1.85}
&& 90.61$\pm$2.17     & 95.98$\pm$1.37     & 90.56$\pm$1.89\\

&MHIM(DSMIL)~\cite{tang2023mhim}    
& 89.87$\pm$2.23            & 92.66$\pm$2.16        & 86.03$\pm$3.25
&& \underline{92.12$\pm$2.59}     & \underline{96.69$\pm$1.26}     & \underline{91.97$\pm$2.44}\\

&R$^2$T-MIL~\cite{tang2024feature}    
& 89.76$\pm$2.39            & 92.89$\pm$1.63        & 85.50$\pm$3.19
&& 91.75$\pm$2.38     & 96.40$\pm$1.13     & 91.26$\pm$2.60\\

&2DMamba~\cite{zhang20252dmamba}    
& 88.18$\pm$1.59            & 90.96$\pm$1.93        & 83.44$\pm$2.03
&& 90.92$\pm$3.06     & 96.27$\pm$1.77  & 90.70$\pm$2.96 \\

& \textbf{MHIM-v2(AB-MIL)}   
& 89.76$\pm$1.81            & 92.77$\pm$1.97        & 85.25$\pm$3.47
&& 91.94$\pm$2.24     & 96.22$\pm$1.44     & 91.89$\pm$2.23\\

& \textbf{MHIM-v2(TransMIL)}   
& \textbf{90.20$\pm$1.49}            & \textbf{93.47$\pm$0.57}        & \textbf{86.47$\pm$2.11}
&& 91.37$\pm$2.04     & 96.29$\pm$1.44     & 91.32$\pm$2.01\\

& \textbf{MHIM-v2(DSMIL)}   
& 89.53$\pm$1.49            & 92.93$\pm$1.86        & 85.66$\pm$1.75
&& \textbf{92.31$\pm$2.62}     & \textbf{96.82$\pm$1.23}     & \textbf{92.29$\pm$2.62}\\

\midrule
\multirow{13}{*}{\rotatebox{90}{\makecell[c]{PLIP}}}
&AB-MIL~\cite{ilse2018attention}  
& 91.87$\pm$1.35            & 94.98$\pm$1.06        & 88.71$\pm$1.68
&& 91.09$\pm$2.35     & 95.59$\pm$1.98     & 91.07$\pm$2.36\\

&CLAM~\cite{clam}        
& 90.87$\pm$2.49            & 95.25$\pm$1.49        & 87.48$\pm$3.16
&& 90.80$\pm$2.35     & 95.46$\pm$1.72     & 90.38$\pm$2.46\\

&DSMIL~\cite{li2021dual}          
& 90.87$\pm$1.44            & 93.70$\pm$1.99        & 87.22$\pm$2.35   
&& 90.70$\pm$3.00     & 95.70$\pm$1.87     & 90.19$\pm$3.11\\

&TransMIL~\cite{shao2021transmil} 
& 90.31$\pm$2.04            & 94.56$\pm$1.28        & 87.07$\pm$2.57
&& 90.52$\pm$3.13     & 95.52$\pm$1.95     & 90.13$\pm$3.31\\

&DTFD-MIL~\cite{zhang2022dtfd}    
& 90.64$\pm$0.83            & 95.39$\pm$1.51        & 87.31$\pm$1.37
&& 90.42$\pm$2.98     & 95.83$\pm$1.75     & 89.91$\pm$3.01\\

&IBMIL~\cite{lin2023interventional}    
& 91.76$\pm$2.27            & 94.66$\pm$1.87        & 88.58$\pm$3.59
&& 91.18$\pm$3.27     & 95.62$\pm$2.09     & 90.94$\pm$3.20\\

&MHIM(AB-MIL)~\cite{tang2023mhim}    
& 90.98$\pm$1.73            & 94.94$\pm$1.56        & 87.80$\pm$1.69
&& 91.74$\pm$1.88     & 96.21$\pm$1.26     & 91.20$\pm$1.89\\

&MHIM(TransMIL)~\cite{tang2023mhim}    
& 90.42$\pm$2.46            & 94.30$\pm$1.72        & 86.91$\pm$3.11
&& 90.80$\pm$2.23     & 96.08$\pm$2.03     & 90.38$\pm$2.49\\

&MHIM(DSMIL)~\cite{tang2023mhim}    
& 90.98$\pm$2.14            & 95.32$\pm$1.59        & 88.02$\pm$2.62
&& 91.27$\pm$2.17     & 96.23$\pm$1.51     & 90.96$\pm$2.36\\

&R$^2$T-MIL~\cite{tang2024feature}    
& 91.87$\pm$1.74            & 95.03$\pm$1.56        & 88.60$\pm$2.60
&& 92.13$\pm$2.55     & 96.40$\pm$1.45     & 91.83$\pm$2.50\\

&2DMamba~\cite{zhang20252dmamba}    
&90.41$\pm$2.94            & 94.77$\pm$2.64        & 87.04$\pm$2.95
&& 90.62$\pm$2.75     & 96.31$\pm$2.07     & 90.29$\pm$2.71 \\

& \textbf{MHIM-v2(AB-MIL)}   
& \underline{91.87$\pm$0.77}            & \underline{95.68$\pm$1.28}        & \underline{88.91$\pm$1.08}
&& \textbf{92.69$\pm$1.43}     & \underline{96.61$\pm$1.09}     & \textbf{92.64$\pm$1.44}\\

& \textbf{MHIM-v2(TransMIL)}   
& 90.31$\pm$2.64            & \textbf{95.80$\pm$1.66}        & 87.53$\pm$3.27
&& 91.27$\pm$2.34     & 96.38$\pm$1.55     & 91.24$\pm$2.34\\

& \textbf{MHIM-v2(DSMIL)}   
& \textbf{92.31$\pm$1.14}            & 95.64$\pm$1.28        & \textbf{89.39$\pm$1.41}
&& \underline{92.60$\pm$1.63}     & \textbf{96.65$\pm$1.23}     & \underline{92.58$\pm$1.64}\\

\bottomrule
\end{tabularx}

\footnotetext{
The highest performance is in bold, and the second-best performance is underlined. The Accuracy and F1-score are determined by the optimal threshold. MHIM (baseline) is the conference version of this paper, and we refer to the improved framework proposed in this paper as MHIM-v2(baseline). 
}
\vspace{-0.3cm}
\end{table*}


\section{Experiments and Results}
\subsection{Datasets and Evaluation Metrics}
\subsubsection{Datasets}
We validate MHIM on various computational pathology tasks, including cancer diagnosis (CAMELYON~\cite{bejnordi2017diagnostic,bandi2018detection}), subtyping (TCGA-NSCLC, TCGA-BRCA), and survival analysis (TCGA-LUAD, TCGA-LUSC, TCGA-BLCA).

\textbf{CAMELYON-16} \cite{bejnordi2017diagnostic} and \textbf{CAMELYON-17} \cite{bandi2018detection} are among the largest publicly available datasets for breast cancer lymph node metastasis diagnosis, both containing binary (metastasis or not) class labels. The CAMELYON-16 dataset includes 270 training WSIs, along with an additional 129 slides as the official test set. Moreover, the CAMELYON-17 dataset encompasses 1000 WSIs from five different medical centers in the Netherlands. Given that the labels for the 500 WSIs in the CAMELYON-17 official test set are not publicly available, we only used the training set of CAMELYON-17, which includes 500 WSIs from 100 cases (with corresponding image-level diagnostic results). We combined CAMELYON-16 and CAMELYON-17 into a single dataset, named \textbf{CAMELYON}, totaling 899 WSIs (591 negative, 308 positive) from 370 cases.

The Non-Small Cell Lung Cancer (\textbf{NSCLC}) project of The Cancer Genome Atlas (TCGA) by the National Cancer Institute (NCI) is the primary dataset for the cancer subtyping task. \textbf{TCGA-NSCLC} is the most common type of lung cancer, accounting for approximately 85\% of all lung cancer cases. This classification includes several subtypes, primarily Lung Adenocarcinoma (\textbf{LUAD}) and Lung Squamous Cell Carcinoma (\textbf{LUSC}). The dataset contains 541 slides from 478 LUAD cases and 512 slides from 478 LUSC cases, with only image-level labels provided. Compared to CAMELYON, the tumor regions in the positive samples of this dataset are significantly larger.

The Breast Invasive Carcinoma (\textbf{TCGA-BRCA}) project is another subtyping dataset we used. TCGA-BRCA includes two subtypes: Invasive Ductal Carcinoma (\textbf{IDC}) and Invasive Lobular Carcinoma (\textbf{ILC}). It contains 779 IDC slides and 198 ILC slides from 1062 cases .

The goal of survival analysis is to estimate the survival probability or survival time of patients over a specific period. Therefore, we used the \textbf{TCGA-LUAD}, \textbf{TCGA-LUSC}, and \textbf{TCGA-BLCA} projects to evaluate the model performance for survival analysis tasks. Unlike the diagnosis and subtyping tasks, the survival analysis datasets are case-based rather than WSI-based. The WSIs of TCGA-LUAD and TCGA-LUSC are identical to those used in the subtyping task but with different annotations. The TCGA-BLCA includes 376 cases of bladder urothelial carcinoma.

\subsubsection{Preprocess}
Following prior works~\cite{clam,shao2021transmil,zhang2022dtfd,tang2024feature}, we crop each WSI into a series of non-overlapping patches of size (256 $\times$ 256) at 20$\times$ magnification and discard the background regions, including holes, as in CLAM~\cite{clam}. Table~\ref{tab:wsi_set} shows the details of the preprocessed datasets, where the average number of patches per dataset is around 10,000. To efficiently handle the large number of patches, we follow the traditional two-stage paradigm, using a pre-trained offline model to extract patch features. This includes a ResNet-50~\cite{he2016deep} pre-trained on ImageNet-1k~\cite{deng2009imagenet}. Specifically, the last convolutional module of the ResNet-50 is removed, and a global average pooling is applied to the final feature maps to generate the initial feature vector. Additionally, we also use state-of-the-art foundation models pre-trained on WSIs, such as PLIP~\cite{huang2023visual} and UNI~\cite{chen2024towards}. The use of these advanced feature extractors enhances the rigor of our experiments and further demonstrates the generalizability of our method.

\subsubsection{Evaluation Metrics}
Following previous work~\cite{shao2021transmil, clam, tang2024feature}, we leverage Accuracy, Area Under Curve (AUC), and F1-score to evaluate model performance in the cancer diagnosis and subtyping tasks. AUC is the primary performance metric in the binary classification task, and we only report AUC in ablation experiments. For the survival analysis task, we refer to related works~\cite{zhou2023cross,chen2021multimodal,zhu2017wsisa} and use the Concordance Index (C-index)~\cite{c-index_harrell} as the evaluation metric. The C-index is considered an extension of AUC that accounts for censored data. This index comprehensively reflects the capability of model to reliably rank survival times based on individual risk scores.


\begin{table}[tb]
\caption{Subtyping results on TCGA-BRCA.}
\label{tab:nsclc}
\centering
\footnotesize
\begin{tabularx}{\linewidth}{Xp{2.cm}ccc}
\toprule
&Methods
            & Accuracy              & AUC                   & F1-score \\ 
\midrule
\multirow{13}{*}{\rotatebox{90}{ResNet-50}}
&AB-MIL     & 86.4$_{\pm4.9}$           & 91.1$_{\pm2.5}$        & 81.6$_{\pm4.7}$\\
&CLAM       & 85.2$_{\pm2.7}$           & 91.7$_{\pm1.8}$        & 80.4$_{\pm3.0}$\\
&DSMIL      & 87.2$_{\pm2.7}$           & 91.6$_{\pm1.3}$        & 82.4$_{\pm2.9}$\\
&TransMIL   & 84.7$_{\pm2.7}$           & 90.8$_{\pm1.9}$        & 79.9$_{\pm2.6}$\\
&DTFD-MIL   & 85.9$_{\pm1.8}$           & 91.4$_{\pm1.6}$        & 81.1$_{\pm2.1}$\\
&IBMIL      & 84.2$_{\pm3.4}$           & 91.0$_{\pm2.3}$        & 79.5$_{\pm3.4}$\\
&MHIM(AB.)  & 86.7$_{\pm5.6}$           & 92.4$_{\pm1.6}$        & 82.4$_{\pm5.5}$\\
&MHIM(Tr.)  & 86.7$_{\pm3.7}$           & 92.6$_{\pm1.7}$        & 82.3$_{\pm3.5}$  \\
&MHIM(DS.)  & 87.4$_{\pm4.1}$           & 92.5$_{\pm1.9}$        & 83.1$_{\pm4.4}$  \\
&R$^2$T-MIL     & 88.3$_{\pm0.7}$           & \textbf{93.2$_{\pm1.5}$}        & 83.7$_{\pm1.0}$  \\
&2DMamba     & 87.53$_{\pm1.7}$           & 92.8$_{\pm2.3}$        &  74.4$_{\pm3.1}$  \\
&MHIM-v2(AB.)   & \underline{88.5$_{\pm3.8}$}           & 92.7$_{\pm1.6}$        & \underline{84.0$_{\pm4.3}$}\\
&MHIM-v2(Tr.)   & \textbf{88.5$_{\pm3.7}$}           & 92.9$_{\pm1.5}$        & \textbf{84.0$_{\pm4.1}$}\\
&MHIM-v2(DS.)   & 87.7$_{\pm2.3}$           & \underline{93.2$_{\pm1.9}$}        & 83.1$_{\pm2.2}$\\
\midrule
\multirow{13}{*}{\rotatebox{90}{PLIP}}
&AB-MIL     & 85.5$_{\pm2.3}$           & 91.7$_{\pm2.3}$        & 80.6$_{\pm2.7}$ \\
&CLAM       & 86.7$_{\pm1.4}$           & 92.2$_{\pm2.0}$        & 81.9$_{\pm1.8}$   \\
&DSMIL      & 85.3$_{\pm3.1}$           & 91.9$_{\pm2.2}$        & 80.3$_{\pm3.0}$ \\
&TransMIL   & 85.8$_{\pm3.4}$           & 92.2$_{\pm2.2}$        & 81.1$_{\pm3.3}$  \\
&DTFD-MIL   & 86.4$_{\pm2.7}$           & 92.2$_{\pm2.4}$        & 81.8$_{\pm2.7}$  \\
&IBMIL      & 87.6$_{\pm1.5}$           & 91.7$_{\pm1.7}$        & 82.8$_{\pm2.0}$  \\
&MHIM(AB.)  & 87.1$_{\pm2.2}$           & 93.2$_{\pm2.0}$        & 82.5$_{\pm2.5}$  \\
&MHIM(Tr.)  & 83.4$_{\pm5.6}$           & 92.9$_{\pm3.1}$        & 79.2$_{\pm5.4}$  \\
&MHIM(DS.)  & 87.1$_{\pm2.6}$           & 92.8$_{\pm2.0}$        & 82.5$_{\pm2.4}$  \\
&R$^2$T-MIL     & 88.8$_{\pm3.2}$           & \textbf{93.8$_{\pm1.2}$}        & \underline{84.6$_{\pm3.6}$}  \\
&2DMamba     & \underline{89.1$_{\pm1.2}$}           & {92.0$_{\pm1.5}$}        & 82.4$_{\pm2.5}$  \\
&MHIM-v2(AB.)   &  \textbf{90.7$_{\pm1.4}$}           & 93.3$_{\pm2.2}$        & \textbf{86.4$_{\pm1.6}$}\\
&MHIM-v2(Tr.)   & 88.3$_{\pm2.8}$           & \underline{93.7$_{\pm2.3}$}        & 83.8$_{\pm3.0}$\\
&MHIM-v2(DS.)   & 88.7$_{\pm2.7}$           & 93.0$_{\pm1.4}$        & 84.2$_{\pm2.3}$\\
\bottomrule
\end{tabularx}
\vspace{-0.3cm}
\end{table}

\subsection{Implementation Details}
Following~\cite{clam,shao2021transmil,zhang2022dtfd}, the offline feature is projected to a 512-dimensional feature vector using a fully-connected layer. The momentum rate of EMA in teacher optimization is set to 0.9999. An Adam optimizer~\cite{kingma2014adam} with a learning rate of $2\times 10^{-4}$ and a weight decay of $1\times 10^{-5}$ is used for model training. The learning rate is adjusted using the cosine annealing strategy. All models were trained for 200 epochs with an early-stopping strategy applied to the cancer diagnosis and subtyping tasks. The patience values for CAMELYON and TCGA are set to 30 and 20, respectively. For survival analysis, all models are trained for 30 epochs. No additional techniques, such as gradient clipping or gradient accumulation, are used to enhance model performance. The batch size is set to 1. All experiments are conducted using 5-fold cross-validation, and the training process does not use a validation set due to data scarcity. We report the mean and standard deviation in the following tables. Regarding computational resources, the cancer diagnosis and subtyping tasks were trained on an NVIDIA RTX 3090, while the survival task was run on an NVIDIA V100 (32G).

\subsection{Main Results}
\subsubsection{Baselines}
We primarily compare our method with well-established approaches such as AB-MIL~\cite{ilse2018attention}, DSMIL~\cite{li2021dual}, CLAM~\cite{clam}, TransMIL~\cite{shao2021transmil}, and DTFD-MIL~\cite{zhang2022dtfd}. AB-MIL (AB.), TransMIL (Tr.), and DSMIL (DS.). Despite their differences in attention mechanisms, these methods can all be considered generalized attention-based MIL methods. We use them as baselines to validate the generalizability of our framework. Additionally, we compare our approach with recent methods such as IBMIL~\cite{lin2023interventional}, MHIM~\cite{tang2023mhim}, R$^2$T-MIL~\cite{tang2024feature}, and 2DMamab~\cite{zhang20252dmamba}, with MHIM being the conference version of this paper. To differentiate, we refer to the improved framework proposed in this paper as MHIM-v2. For consistency, the results of all other methods are reproduced using their official codes under the same experimental settings.

\subsubsection{Cancer Diagnosis and Subtyping}
As shown in the left part of Table~\ref{tab:main_comp}, the performance of nearly all models on the newly constructed CAMELYON dataset is suboptimal. Some models, such as DSMIL~\cite{li2021dual} and TransMIL~\cite{shao2021transmil}, were only validated on the CAMELYON-16 dataset in their original papers, resulting in inferior performance on the larger CAMELYON dataset compared to classic models like ABMIL. However, our proposed MHIM-v2 framework demonstrates superior performance on such a challenging dataset, and this advantage is not just limited to a specific baseline model. MHIM-v2 achieves high-level performances across three baseline models. Specifically, MHIM-v2 (Trans.) improves the AUC by 0.7\% compared to the second-best method named R$^2$T-MIL. Additionally, due to inaccuracies in hard instance mining and the omission of critical features, MHIM-MIL performs poorly with PLIP features, achieving an AUC of only 94.94\%, which is slightly lower than the baseline model ABMIL. In contrast, MHIM-v2 shows significant improvement with PLIP features, increasing the AUC by 0.7\%, thus validating the effectiveness of our improvements over MHIM.

As shown in Tables~\ref{tab:main_comp} and~\ref{tab:nsclc}, the cancer subtyping results on the TCGA-NSCLC and TCGA-BRCA datasets exhibit similar phenomena. The SOTA R$^2$T-MIL model achieves a slightly better AUC on the BRCA compared to MHIM-v2. However, experiments demonstrate that MHIM-v2 outperforms R$^2$T-MIL in terms of F1-score and Accuracy, indicating its performance advantages. More importantly, when MHIM-v2 is applied to more advanced models like TransMIL and DSMIL, it generally achieves better performance, even if the baselines may not perform as well as the classic ABMIL. We attribute this to the fact that more advanced and complex network architectures may not fully exploit their learning capabilities due to data limitations. In contrast, MHIM-v2 effectively enhances the learning of these more complex networks on limited WSIs through high-quality hard instance mining, thereby maximizing their potential in cancer subtyping tasks.

\begin{table}[tb]
\caption{\justifying Survival analysis results on three main datasets.}
\label{tab:survival_prediction}
\centering
\footnotesize
\begin{tabularx}{\linewidth}{Xp{2.cm}ccc}
\toprule
& Methods & BLCA & LUAD & LUSC \\ 
\midrule
\multirow{13}{*}{\rotatebox{90}{ResNet-50}}
& AB-MIL     & 59.0$_{\pm5.6}$     & 59.5$_{\pm4.5}$     & 59.4$_{\pm7.1}$\\
& CLAM       & 59.0$_{\pm3.5}$     & 59.2$_{\pm4.1}$     & 60.0$_{\pm2.9}$\\
& DSMIL      & 59.2$_{\pm4.2}$     & 59.6$_{\pm4.9}$     & 59.3$_{\pm4.5}$\\
& TransMIL   & 60.4$_{\pm2.8}$     & 62.2$_{\pm1.0}$     & 59.0$_{\pm4.6}$\\
& DTFD-MIL   & 59.3$_{\pm4.4}$     & 60.1$_{\pm3.5}$     & 59.0$_{\pm4.1}$\\
& IBMIL      & 59.1$_{\pm5.3}$     & 59.8$_{\pm4.1}$     & 58.8$_{\pm3.3}$\\
& MHIM(AB.)   & 59.6$_{\pm4.2}$     & 59.9$_{\pm3.6}$     & 59.7$_{\pm3.3}$\\
& MHIM(Tr.)   & 60.6$_{\pm1.4}$     & 62.5$_{\pm2.4}$     & 60.9$_{\pm5.3}$\\
& MHIM(DS.)   & 59.4$_{\pm3.9}$     & 60.7$_{\pm5.0}$     & 60.1$_{\pm4.1}$\\
& R$^2$T-MIL & 61.0$_{\pm2.1}$     & \underline{64.0$_{\pm4.4}$}     & 61.0$_{\pm6.9}$\\
& 2DMamba & \underline{61.2$_{\pm3.9}$}     & 61.1$_{\pm4.3}$     & 59.02$_{\pm4.1}$\\
& MHIM-v2(AB.)    & 60.8$_{\pm4.0}$     & 60.0$_{\pm4.9}$     & \underline{61.0$_{\pm3.9}$}\\
& MHIM-v2(Tr.)    & \textbf{61.9$_{\pm3.7}$}     & \textbf{65.1$_{\pm4.2}$}     & \textbf{62.7$_{\pm3.8}$}\\
& MHIM-v2(DS.)    & 59.7$_{\pm3.3}$     & 60.4$_{\pm4.2}$     & 60.3$_{\pm5.2}$\\
\midrule
\multirow{13}{*}{\rotatebox{90}{UNI}}
& AB-MIL     & 59.7$_{\pm5.8}$     & 65.1$_{\pm5.1}$     & 58.9$_{\pm4.1}$\\
& CLAM       & 59.5$_{\pm2.9}$     & 64.9$_{\pm3.1}$     & 59.9$_{\pm3.1}$\\
& DSMIL      & 60.1$_{\pm4.0}$     & 63.8$_{\pm2.5}$     & 61.5$_{\pm4.7}$\\
& TransMIL   & 61.2$_{\pm2.9}$     & 66.5$_{\pm1.8}$     & 61.1$_{\pm2.5}$\\
& DTFD-MIL   & 61.0$_{\pm5.9}$     & 65.0$_{\pm2.3}$     & 60.8$_{\pm6.0}$\\
& IBMIL      & 58.6$_{\pm5.5}$     & 63.3$_{\pm2.8}$     & 60.3$_{\pm4.8}$\\
& MHIM(AB.)   & 60.2$_{\pm4.4}$    & 66.3$_{\pm7.0}$     & 60.9$_{\pm4.6}$\\
& MHIM(Tr.)   & 62.1$_{\pm0.7}$    & 66.9$_{\pm1.5}$     & 62.0$_{\pm4.4}$\\
& MHIM(DS.)   & 61.5$_{\pm4.3}$    & 66.4$_{\pm2.5}$     & 62.1$_{\pm2.3}$\\
& R$^2$T-MIL & 61.5$_{\pm4.3}$     & 66.4$_{\pm2.6}$     & 62.6$_{\pm4.2}$\\
& 2DMamba & \textbf{64.8$_{\pm4.3}$}     & 63.8$_{\pm7.6}$     & 61.9$_{\pm5.9}$\\
& MHIM-v2(AB.)    & 61.9$_{\pm5.5}$     & \underline{67.7$_{\pm4.5}$}     & 61.3$_{\pm4.5}$\\
& MHIM-v2(Tr.)    & \underline{63.0$_{\pm1.6}$}     & \textbf{67.8$_{\pm2.2}$}     & \underline{62.7$_{\pm2.9}$}\\
& MHIM-v2(DS.)    & 61.9$_{\pm5.8}$     & 67.4$_{\pm3.3}$     & \textbf{63.1$_{\pm5.3}$}\\
\bottomrule
\end{tabularx}
\end{table}

\begin{table}[tb]
\centering
\caption{\justifying Performance of different methods on cross-source validation using the CPTAC dataset.}
\label{tab:cross-source}
\footnotesize
\begin{tabularx}{\linewidth}{p{2.cm}cccc}
    \toprule
    {\multirow{2}{*}{Methods}}   
    & \multicolumn{2}{c}{CPTAC$_{\textrm{NSCLC}}$} 
    & \multicolumn{2}{c}{CPTAC$_{\textrm{LUAD}}$} \\ 
    \cmidrule(lr){2-3} \cmidrule(lr){4-5} 
    & R50         & PLIP        & R50     & UNI\\ 
    \midrule
    CLAM                       & 74.9$_{\pm1}$ & 80.6$_{\pm1}$ & 47.1$_{\pm2}$ & 55.6$_{\pm3}$ \\
    R$^2$T-MIL                & 76.3$_{\pm1}$ & 81.4$_{\pm1}$ & 51.4$_{\pm2}$ & 57.0$_{\pm4}$ \\
    \midrule
    \textit{AB-MIL}           & 74.1$_{\pm1}$ & 81.4$_{\pm2}$ & 48.6$_{\pm1}$ & 52.0$_{\pm4}$ \\
    MHIM(AB.)                 & 76.5$_{\textcolor{red}{+2}}$ & 81.2$_{\textcolor{blue}{-.1}}$ & 49.1$_{\textcolor{red}{+.5}}$ & 55.8$_{\textcolor{red}{+4}}$ \\
    MHIM-v2(AB.)              & 76.7$_{\textcolor{red}{+3}}$ & 82.3$_{\textcolor{red}{+.9}}$ & 49.1$_{\textcolor{red}{+.6}}$ & 56.7$_{\textcolor{red}{+5}}$ \\
    \midrule
    \textit{TransMIL}         & 77.2$_{\pm2}$ & 81.2$_{\pm1}$ & 50.4$_{\pm9}$ & 53.3$_{\pm3}$ \\
    MHIM(Tr.)                 & 79.0$_{\textcolor{red}{+2}}$ & 82.7$_{\textcolor{red}{+2}}$ & 54.0$_{\textcolor{red}{+3}}$ & 57.0$_{\textcolor{red}{+4}}$ \\
    MHIM-v2(Tr.)              & \textbf{79.6$_{\textcolor{red}{+3}}$} & \textbf{83.1$_{\textcolor{red}{+2}}$} & \textbf{54.9$_{\textcolor{red}{+4}}$} & \textbf{57.6$_{\textcolor{red}{+5}}$} \\
    \bottomrule
\end{tabularx}
\footnotetext{
 All methods are trained on TCGA datasets.  Results for the NSCLC were reported as AUC, while those for the LUAD were reported as C-index. We present the \textcolor{red}{increase} and \textcolor{blue}{decrease} in performance of our methods compared to the baseline, and the standard deviations of other methods.
}
\end{table}

\begin{table*}[t]
\centering
\caption{\justifying Comparison of computational efficiency of different MIL methods. We report the model size (Parameter), the training time per epoch (Train Time), the peak memory usage (Memory), the inference speed on the CAMELYON dataset, and performances on all three tasks. AB. and Trans. denote AB-MIL and TransMIL baselines, respectively.}
\label{tab:effi}
\footnotesize
\begin{tabularx}{\textwidth}{lccccccc}
\toprule
Model & Parameter$\downarrow$ &Train Time$\downarrow$ & Memory$\downarrow$ & Inference Speed$\uparrow$ &CAMELYON$\uparrow$ & NSCLC$\uparrow$  & LUAD$\uparrow$  \\ \midrule
\textit{ABMIL}        & 0.66M & 3.1s &2.3G & 1250 & 91.59  & 95.32 & 59.47 \\ 
DTFD-MIL          &0.99M & 5.1s &2.1G & 325 & 91.49  & 95.55 & 60.05 \\
R$^2$T-MIL &2.70M & 6.5s &10.1G & 236 & 92.73  & \textbf{96.40} & \underline{64.01} \\ 
MHIM(AB.) &0.65M & 4.1s &2.2G & 1250 & 92.30  & 96.02 & 59.87 \\
MHIM-v2(AB.) &1.64M & 4.3s &2.8G & 714 & \underline{92.77}  & 96.22 & 59.98  \\
\midrule
\textit{TransMIL}       & 2.67M & 13.2s & 10.6G & 76 & 91.23  & 94.97 & 62.15\\
MHIM(Trans.) &2.67M & 10.1s &5.5G & 76 & 92.62  &  95.98 & 62.52\\
MHIM-v2(Trans.) &3.72M & 10.5s &5.6G & 72 & \textbf{93.47}  & \underline{96.29} & \textbf{65.11} \\
\bottomrule
\end{tabularx}
\footnotetext{
 We report the model size (Parameter), the training time per epoch (Train Time), the peak memory usage (Memory), the inference speed on the CAMELYON dataset, and performances on all three tasks. AB. and Trans. denote AB-MIL and TransMIL baselines, respectively.
}
\vspace{-0.3cm}
\end{table*}

\subsubsection{Survival Analysis}
Table~\ref{tab:survival_prediction} presents the experimental results for three survival analysis datasets. Notably, the proposed MHIM-v2 achieved favorable results across all three datasets. Specifically, leveraging the SOTA large pathology model UNI, MHIM-v2 improved the unimodal performance for BLCA, LUAD, and LUSC to 63.04\%, 67.70\%, and 63.13\%, respectively, surpassing the previous SOTA model R$^2$T-MIL. More importantly, compared to the conference version, MHIM-v2 showed performance improvements across different datasets and baselines, such as a +1.27\% increase on the BLCA-R50 TransMIL baseline and a +0.63\% increase on the LUAD-R50 AB-MIL baseline.  These enhancements enabled MHIM-v2 to outperform R$^2$T-MIL, demonstrating superior performance. Furthermore, with high-quality features from UNI, both the MHIM-v2 and its conference version demonstrated a more significant advantage over existing methods. This underscores the excellent adaptability of the MHIM framework to advanced features from large pathology models.

\subsubsection{Cross-source Validation}
We evaluated the transferability and generalization of our model by applying it to different datasets for validation, as shown in Table~\ref{tab:cross-source}. Specifically, the model trained on the TCGA-NSCLC dataset was applied to the CPTAC-NSCLC dataset for the subtyping task. Similarly, the model trained on the TCGA-LUAD dataset was validated on the CPTAC-LUAD dataset for survival analysis. In both cases, the CPTAC datasets were used solely as test sets, dedicated to these validation tasks. This cross-source validation demonstrates strong transferability and generalization of our model across diverse data sources. The results indicate its potential for broader applicability and confirm its ability to adapt effectively to new, unseen datasets.

\begin{table}[tbp]
\footnotesize
\setlength{\tabcolsep}{2.3mm}
\centering
\caption{\justifying Comparison of different modules in MHIM-v2.}
\label{tab:abl}
\begin{tabularx}{\linewidth}{p{0.1cm}p{2.3cm}cccc}
\toprule
&{\multirow{2}{*}{Module}} & \multicolumn{2}{c}{CAMELYON} & \multicolumn{2}{c}{NSCLC} \\ \cmidrule(lr){3-4} \cmidrule(lr){5-6}
 && AB.         & Trans.        & AB.     & Trans.     \\\midrule
\multirow{5}{*}{\rotatebox{90}{R50}}
&Baseline                  & 91.6         & 91.2          & 95.3     & 95.0        \\
&+RHSM                   & 92.1         & 92.4           & 95.9     & 95.1 \\
&+RHSM+GRN                     & 92.5         & 92.4           & 96.1     & 95.8        \\
&+RHSM+CL                     & 92.3         & 92.6           & 95.9     & 95.5        \\
&+RHSM+GRN+CL                  & \textbf{92.8}         & \textbf{93.5}           & \textbf{96.2}     & \textbf{96.3} \\   \midrule
\multirow{5}{*}{\rotatebox{90}{PLIP}}
&Baseline                  & 95.0         & 94.6           & 95.6     & 95.5        \\
&+RHSM                    & 95.2         & 93.9          & 96.3     & 95.9 \\
&+RHSM+GRN                     & 95.2         & 94.9           & 96.4     & 96.0        \\
&+RHSM+CL                     & 95.2         & 94.4           & 96.3     & 95.8        \\
&+RHSM+GRN+CL                  & \textbf{95.7}         & \textbf{95.8}           & \textbf{96.6}     & \textbf{96.4} \\ 
\bottomrule 
\end{tabularx}
\footnotetext{
RHSM denotes the Randomly High Score Masking. GRN refers the Global Recycle Network. CL is the Consistency Loss.
}
\end{table}

\subsection{Computational Cost Analysis}
Here, we report the training time and GPU memory requirements of different MIL models on an NVIDIA RTX-3090 GPU. The upper part of Table~\ref{tab:effi} compares several MIL frameworks using ABMIL~\cite{ilse2018attention} as a baseline. The results indicate that traditional MIL frameworks often introduce additional parameters because of their complex structures, leading to reduced efficiency. For example, the parameter size of DTFD-MIL~\cite{zhang2022dtfd} nearly 1.5 times that of ABMIL (from 657K to 987K), increasing training time by 30\%, yet its performance is suboptimal. In contrast, although MHIM-v2 introduced approximately 1M more parameters than MHIM-MIL and their baseline, both MHIM-MIL and MHIM-v2 achieve significant performance improvements by incorporating momentum teachers with almost no increase in computational cost.  Existing Transformer-based MIL methods generally incur high computational costs, primarily because of their extensive parameter counts and self-attention mechanisms.
For instance, TransMIL~\cite{shao2021transmil}, which first applied a pure Transformer structure to this area, has 4 times the parameters of ABMIL, 3 times the training time, and nearly 4.5 times the memory consumption. Although R$^2$T-MIL employs regional MSA to mitigate the efficiency impact of the Transformer, its computational cost remains significantly higher than non-Transformer models. By using a high ratio masking strategy, the MHIM-v2 framework significantly reduces the computational cost of such methods (training time decreased by 24\%, memory usage reduced by 48\%) and enhances stability (AUC standard deviation on CAMELYON is 0.57\%). Compared to MHIM-MIL, MHIM-v2 incurs a slight increase in computational cost due to the introduction of a recovery network. However, this cost is acceptable. Compared to MHIM-MIL (TransMIL), the training time for MHIM-v2 increases by only 0.4 seconds, memory usage by 0.1 GB, and the number of images processed per second decreases by 4. However, this minor cost results in a significant performance improvement.


\subsection{Ablation Study}

\subsubsection{Important Modules}
Table~\ref{tab:abl} presents the ablation study of different modules in MHIM-v2 on two datasets. We first introduce the Randomly High Score Masking (RHSM) strategy, which employs a teacher model to mine hard instances during training. This strategy enhances the AUC of both MIL models on the CAMELYON dataset by 0.5\% and 1.1\%, respectively, indicating that the incorporation of hard instance mining during training aids in constructing more accurate classification boundaries for mainstream MIL models. A detailed discussion on the RHSM strategy will be provided in Section~\ref{sec:diff_ptp}, and a discussion on choosing the teacher model will be given in Section~\ref{sec:diff_tea}.
Apart from the RHSM strategy, comparing rows 4 and 5 reveals that the Global Recycle Network (GRN) and High Ratio Masking reduce redundancy for the student model and provide more diverse hard-instances. 
These improves the AUC of the baseline models on the TCGA-NSCLC dataset by 0.3\% and 0.8\%, respectively, confirming the positive impacts of these techniques on student model training. 
A more detailed discussion of GRN will be provided in Section~\ref{sec:merge}.
After incorporating the Consistency Loss (CL) into the objective function, the complete MHIM-v2 framework achieves the best AUC performance of 95.8\% and 96.6\% on the CAMELYON and TCGA-NSCLC datasets, respectively. For subsequent ablation experiments, we include consistency loss by default to facilitate the optimization of our framework.

\begin{table}[tb]
\footnotesize
\centering
\caption{\justifying Comparison between different easy instance assessment strategies.}
\label{tab:pt}
\begin{tabularx}{\linewidth}{p{0.1cm}p{2.5cm}cccc}
    \toprule
    & {\multirow{2}{*}{Strategy}} 
    & \multicolumn{2}{c}{CAMELYON} 
    & \multicolumn{2}{c}{NSCLC} \\
    \cmidrule(lr){3-4} \cmidrule(lr){5-6}
    && AB. & Trans. & AB. & Trans. \\
    \midrule
    \multirow{3}{*}{\rotatebox{90}{R50}}
    & Baseline          & 91.6  & 91.2  & 95.3  & 95.0 \\
    & Attention        & 92.2  & 92.4  & 96.1  & 95.4 \\
    & Instance Probability & \textbf{92.8}  & \textbf{93.5}  & \textbf{96.2}  & \textbf{96.3} \\
    \midrule
    \multirow{3}{*}{\rotatebox{90}{PLIP}}
    & Baseline          & 95.0  & 94.6  & 95.6  & 95.5 \\
    & Attention        & 95.5  & 95.2  & 96.4  & 96.1 \\
    & Instance Probability & \textbf{95.7}  & \textbf{95.8}  & \textbf{96.6}  & \textbf{96.4} \\
    \bottomrule
\end{tabularx}
\footnotetext{
The proposed class-aware instance probability shows advantages over class-agnostic attention on multiple benchmarks.
}
\end{table}

\subsubsection{Class-aware Instance Probability}
\label{sec:diff_ptp}
We introduce class-aware instance probability instead of attention scores in MHIM-MIL to more accurately mine hard instances. To validate the effectiveness of this improvement, ablation experiments were conducted. Table~\ref{tab:pt} shows that, although attention scores can mine some hard instances, their accuracy is insufficient, resulting in limited performance improvement for the baseline model. MHIM-v2 utilizes instance probability to select more challenging instance features for training the student model, thereby enhancing training effectiveness. While this improvement generally enhances model performance across various settings, we found it especially beneficial for the more complex TransMIL, which can better model classification boundaries from hard instances. Specifically, on the CAMELYON using R50 features, the TransMIL achieved a 1.1\% AUC improvement due to this enhancement. Clearly, this experiment demonstrates the superior hard instance mining capability of instance probability compared to attention score.

\begin{table}[tb]
\footnotesize
\centering
\caption{\justifying Ablation of different instance mining methods.}
\label{tab:mask}
\begin{tabularx}{\linewidth}{p{1.7cm}ccccc}
    \toprule
    {\multirow{2}{*}{Strategy}} 
    & {\multirow{2}{*}{Memo.}} 
    & \multicolumn{2}{c}{CAMELYON} 
    & \multicolumn{2}{c}{NSCLC} \\
    \cmidrule(lr){3-4} \cmidrule(lr){5-6}
    && R50 & PLIP & R50 & PLIP \\
    \midrule
    \textit{ABMIL}    & 2.3G  & 91.6  & 95.0  & 95.3  & 95.6 \\
    RHSM+CL           & 2.7G  & 92.3  & 95.2  & 95.9  & 96.3 \\
    +RSM              & 2.6G  & 92.6  & 95.4  & 96.2  & 96.5 \\
    +LSM+GRN          & 2.8G  & 92.3  & 95.4  & 96.0  & 96.4 \\
    +RSM+GRN          & 2.8G  & \textbf{92.8} & \textbf{95.7} & \textbf{96.2} & \textbf{96.6} \\
    \midrule
    \textit{TransMIL} & 10.6G & 91.2  & 94.6  & 95.0  & 95.5 \\
    RHSM+CL           & 10.6G & 92.6  & 94.4  & 95.5  & 95.8 \\
    +RSM              & 5.5G  & 93.0  & 95.1  & 95.6  & 96.3 \\
    +LSM+GRN          & 5.6G  & 92.5  & 95.2  & 95.8  & 96.0 \\
    +RSM+GRN          & 5.6G  & \textbf{93.5} & \textbf{95.8} & \textbf{96.3} & \textbf{96.4} \\
    \bottomrule
\end{tabularx}
\footnotetext{
Based on RHSM, we ablated two masking strategies and investigated the impact of the proposed GRN. We found that RSM not only shows advantages in performance but also significantly reduces memory consumption. Meanwhile, GRN alleviates the critical information loss problem of large-scale masking at a very low cost.
}
\end{table}

\subsubsection{Global Recycle Network}
\label{sec:merge}
We introduce a large-ratio Random-Score Masking (RSM) module within the Global Recycle Network (GRN) as an enhancement to the MHIM-MIL. This module simplifies the hard instance mining strategy and significantly reduces the computational cost of the baselines while enhancing the student model's utilization of key features. The results in Table~\ref{tab:mask} confirm this improvement. By comparing the second and third rows of the table, it is evident that the RSM not only reduces computational costs but also improves model performance. This reveals the presence of significant redundancy and substantial noise in the WSI. However, we believe that this random masking may lead to the omission of critical instance features. This issue is particularly pronounced in the CAMELYON, where key features are sparse. Consequently, the GRN achieves the best AUC improvements on the CAMELYON with PLIP features, showing increases of 0.3\% and 0.7\%. These improvements do not incur high computational costs, demonstrating the efficiency of the network. Additionally, the results in the fourth row indicate that the Low-Score Masking (LSM) strategy with the same masking ratio does not yield satisfactory results. This highlights the superiority of RSM in providing high-quality, diverse hard instances.

\begin{table}[tbp]
\footnotesize
\centering
\caption{Comparison of different types of teachers. }
\label{tab:diff_tea}
\begin{tabularx}{\linewidth}{p{0.1cm}p{2.4cm}cccc}
    \toprule
    & {\multirow{2}{*}{Teacher}} 
    & \multicolumn{2}{c}{CAMELYON} 
    & \multicolumn{2}{c}{NSCLC} \\
    \cmidrule(lr){3-4} \cmidrule(lr){5-6}
    && AB. & Trans. & AB. & Trans. \\
    \midrule
    \multirow{5}{*}{\rotatebox{90}{R50}}
    & Baseline       & 91.6  & 91.2  & 95.3  & 95.0 \\
    & Student copy   & 92.5  & 91.7  & 94.1  & 96.0 \\
    & Init.          & 92.3  & 92.8  & 96.2  & 95.9 \\
    & Momentum       & 92.4  & 92.5  & 95.9  & 96.1 \\
    & Init.+Momentum & \textbf{92.8} & \textbf{93.5} & \textbf{96.2} & \textbf{96.3} \\
    \midrule
    \multirow{5}{*}{\rotatebox{90}{PLIP}}
    & Baseline       & 95.0  & 94.6  & 95.6  & 95.5 \\
    & Student copy   & 95.2  & 94.6  & 96.1  & 95.9 \\
    & Init.          & 95.1  & 94.7  & 96.4  & 96.0 \\
    & Momentum       & 95.3  & 94.6  & 96.3  & 96.0 \\
    & Init.+Momentum & \textbf{95.7} & \textbf{95.8} & \textbf{96.6} & \textbf{96.4} \\
    \bottomrule
\end{tabularx}
\centering
\includegraphics[width=0.7\linewidth]{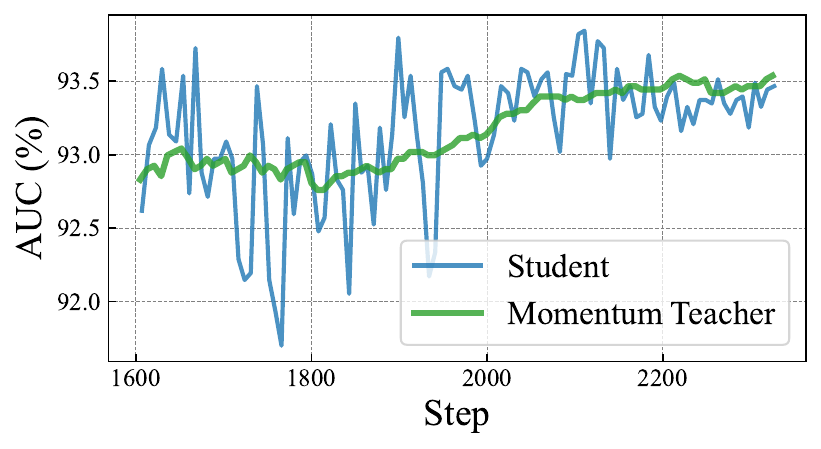}


\footnotetext{\justifying
Momentum denotes the teacher is updated by EMA strategy. Init. indicates the initialization of the teacher with pre-trained baseline model. The bottom figure compares the stability of the momentum teacher and the non-batch gradient updated student during training.
}
\vspace{-0.3cm}
\end{table}

\begin{table}[tb]
\centering
\caption{\justifying Comparison of different MIL models under student initialization.}
\label{tab:init_stu}
\footnotesize
\begin{tabularx}{\linewidth}{p{2.cm}cccc}
    \toprule
    {\multirow{2}{*}{Model}} 
    & \multicolumn{2}{c}{CAMELYON} 
    & \multicolumn{2}{c}{NSCLC} \\
    \cmidrule(lr){2-3} \cmidrule(lr){4-5}
    & R50 & PLIP & R50 & PLIP \\
    \midrule
    \multicolumn{2}{l}{\textit{w/ student init.}} &&& \\
    AB-MIL          & 91.7$_{\textcolor{red}{+.1}}$  & 95.0$_{\textcolor{red}{+.0}}$  & 95.2$_{\textcolor{blue}{-.1}}$  & 95.5$_{\textcolor{blue}{-.2}}$ \\
    MHIM-v2(AB.)    & \textbf{92.8}$_{\textcolor{red}{+.9}}$  & \textbf{95.7}$_{\textcolor{red}{+.4}}$  & \textbf{96.2}$_{\textcolor{red}{+.1}}$  & \textbf{96.6}$_{\textcolor{red}{+.4}}$ \\
    TransMIL        & 91.8$_{\textcolor{red}{+.5}}$  & 94.5$_{\textcolor{blue}{-.1}}$  & 94.5$_{\textcolor{blue}{-.5}}$  & 95.4$_{\textcolor{blue}{-.5}}$ \\
    MHIM-v2(Tr.)    & \textbf{93.5}$_{\textcolor{red}{+1.}}$  & \textbf{95.8}$_{\textcolor{red}{+1.}}$  & \textbf{96.3}$_{\textcolor{red}{+.6}}$  & \textbf{96.4}$_{\textcolor{red}{+.7}}$ \\
    \midrule
    \multicolumn{2}{l}{\textit{w/ student init.}} &&& \\
    CLAM-SB         & 91.4$_{\textcolor{blue}{-.1}}$  & 95.0$_{\textcolor{blue}{-.3}}$  & 95.5$_{\textcolor{red}{+.1}}$  & 95.4$_{\textcolor{blue}{-.1}}$ \\
    DTFD-MIL        & 91.7$_{\textcolor{red}{+.2}}$  & 95.5$_{\textcolor{red}{+.2}}$  & 95.4$_{\textcolor{blue}{-.2}}$  & 95.9$_{\textcolor{red}{+.1}}$ \\
    R$^2$T-MIL      & 92.3$_{\textcolor{blue}{-.6}}$  & 95.1$_{\textcolor{red}{+.1}}$  & 96.4$_{\textcolor{blue}{-.0}}$  & 95.9$_{\textcolor{blue}{-.5}}$ \\
    \bottomrule
\end{tabularx}
\centering
\includegraphics[width=0.7\linewidth]{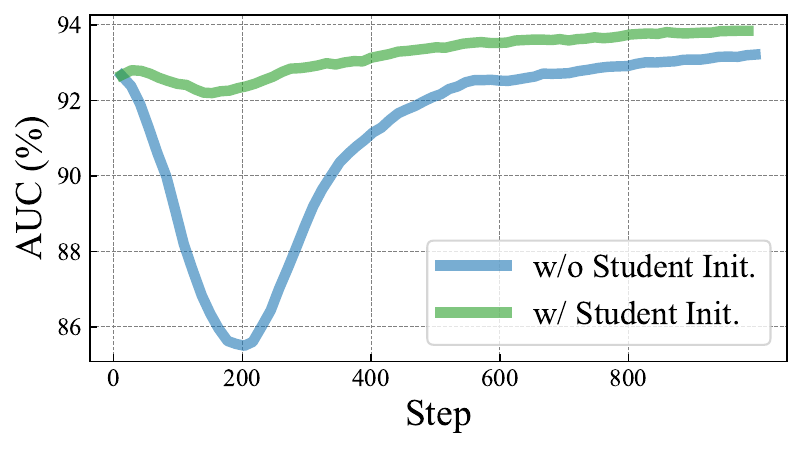}
\footnotetext{\justifying
The \textcolor{red}{red} and \textcolor{blue}{blue} values represent the performance \textcolor{red}{increase} or \textcolor{blue}{decrease} of the algorithm relative to the methods without initialization.
The bottom figure compares the performance of teacher models under initialized or uninitialized student.
}

\end{table}

\subsubsection{Model Choice of MHIM}
\noindent\textbf{Choice of Teacher Model. }
\label{sec:diff_tea}
In MHIM-v2, we employ a Teacher model to mine hard instances and facilitate the training of the Student model. In Table~\ref{tab:diff_tea}, we comprehensively investigate the effects of various choices of Teacher networks. We initially adopts a simple structure where the teacher and student models share parameters. The Student conducts masked hard instance mining prior to training. Due to the non-batch gradient update, the unstable performance of the Student model makes the strategy susceptible to noise, resulting in sub-optimal performance. Second, we adopt a momentum teacher, which shares the same network structure as the Student model and is updated with the EMA strategy. This updating strategy enhances the stability of momentum teachers, as shown in the figure below Table~~\ref{tab:diff_tea}, and enables MHIM-v2 to achieve up to 1.74\% and 0.77\% performance improvement under the two baselines, respectively. After initialization with a pre-trained baseline, the momentum teacher achieves the best performance. However, a fixed teacher fails to learn new knowledge, emphasizing the significance of iterative optimization.

\textbf{Initialization of Student Model. }
We initialize the fully connected layer of the student network with a pre-trained baseline model to reduce the risk of collapse from the Siamese structure, as detailed in~\cite{caron2021dino}. The figure below Table~\ref{tab:init_stu} illustrates how this initialization affects the performance of the teacher model. An uninitialized student model exhibits slow initial training, which impairs the performance of the teacher model and hinders the iterative optimization of the framework. Table~\ref{tab:init_stu} tabulates the performances of different MIL approaches with initialization, where the {red} and {blue} values represent the performance {increase} or {decrease} relative to their uninitialized versions.
The upper part of Table~\ref{tab:init_stu} shows a significant difference in the final performance of the student model with and without this initialization. Additionally, we applied the same initialization to mainstream MIL models to investigate whether it boosts performance by aiding Siamese structure optimization.
The lower part of Table~\ref{tab:init_stu} indicates that this initialization does not significantly enhance the performance of existing mainstream MIL models and sometimes even degrades it. Our experiments confirm that initializing the first fully connected layer of the student facilitates the iterative optimization of the MHIM framework, rather than serving as a universal trick for improving MIL model performance.

\begin{table}[tb]
\centering
\footnotesize
\caption{\justifying Ablation study on error masking mitigation techniques.}
\label{tab:decay}
\begin{tabular}{lcccc}
\toprule
    {\multirow{2}{*}{Model}} 
    & \multicolumn{2}{c}{CAMELYON} 
    & \multicolumn{2}{c}{NSCLC} \\
    \cmidrule(lr){2-3} \cmidrule(lr){4-5}
    & R50 & PLIP & R50 & PLIP \\
\midrule
MHIM-v2(AB.)        & 92.8  & 95.7  & 96.2  & 96.6  \\
\quad w/o Decay       & 92.7 & 95.2 & 96.2  & 96.5 \\
\quad w/o RHSM          & 92.0 & 95.0 & 96.0 & 96.4  \\
MHIM-v2(Tr.)        & 93.5  & 95.8  & 96.3  & 96.4  \\
\quad w/o Decay       & 92.9 & 94.4 & 95.9 & 95.9 \\
\quad w/o RHSM          & 92.8 & 94.7 & 95.9 & 96.2  \\
\bottomrule
\end{tabular}
\footnotetext{\justifying
We employ a decaying high-score masking ratio (Decay) and a randomly high-score masking policy (RHSM) to minimize the risk of removing informative regions.
}
\end{table}

\begin{figure}[tb]
\centering
    \includegraphics[width=\linewidth]{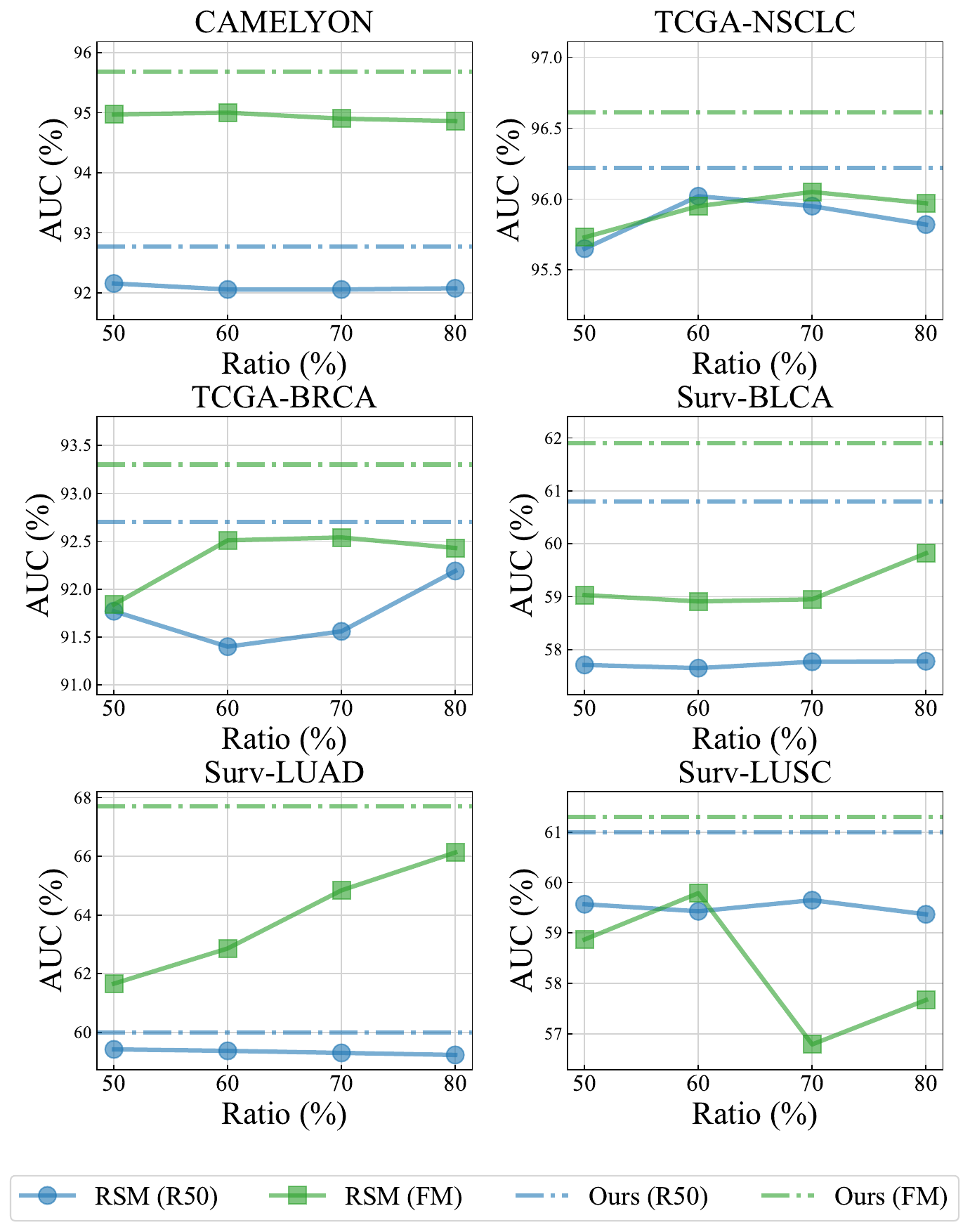}
    \caption{Performance comparison between the simple Random Masking Strategy (RSM) and MHIM-v2 (Ours).}
    \label{fig:hsm}
\end{figure}

\begin{figure}[tb]
\centering
    \includegraphics[width=\linewidth]{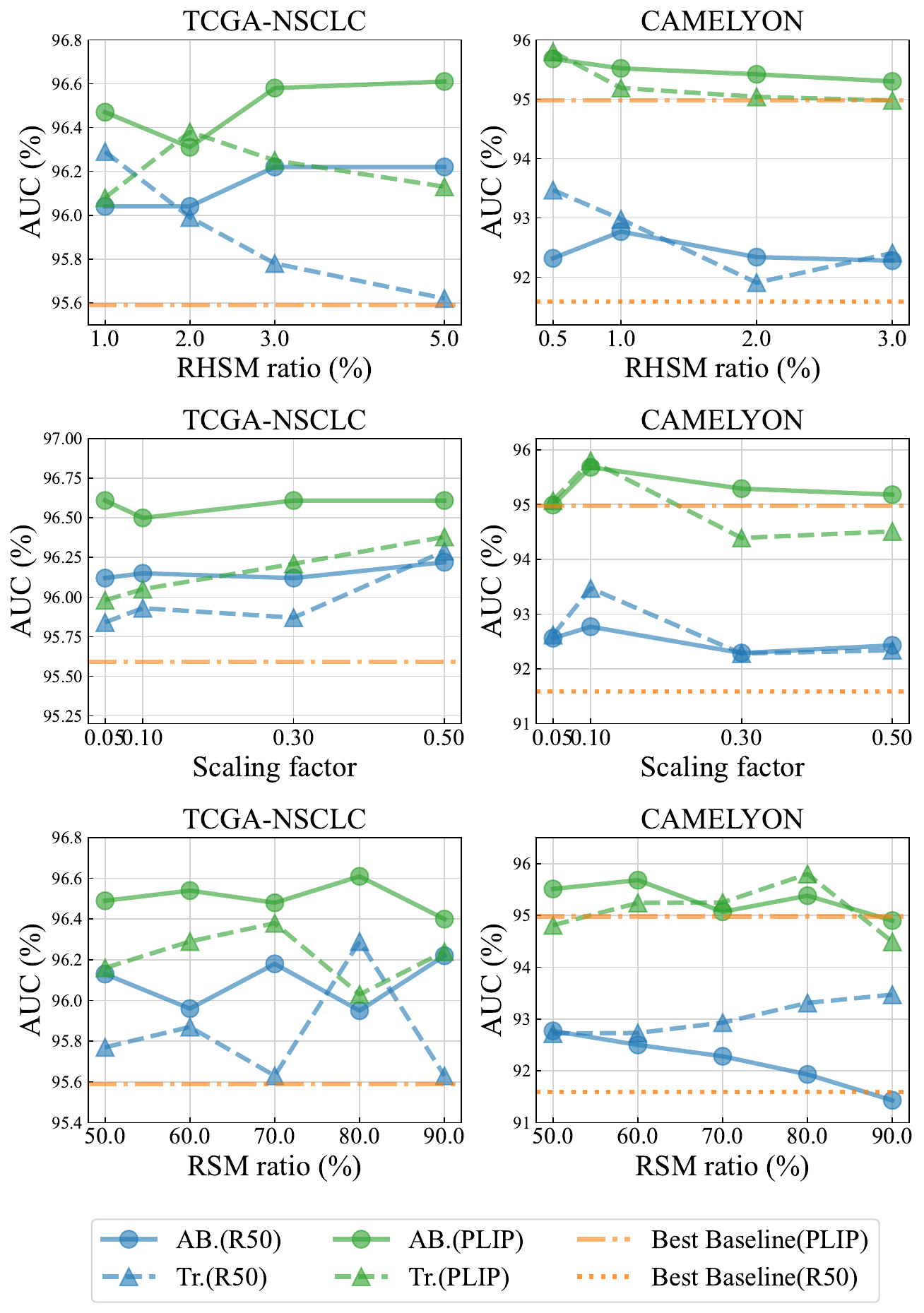}
    \caption{The performances of MHIM-v2 under different important hyperparameters.}
    \label{fig:hyper-para}
\end{figure}





\begin{figure*}[t!]
\centering
    \includegraphics[width=0.75\linewidth]{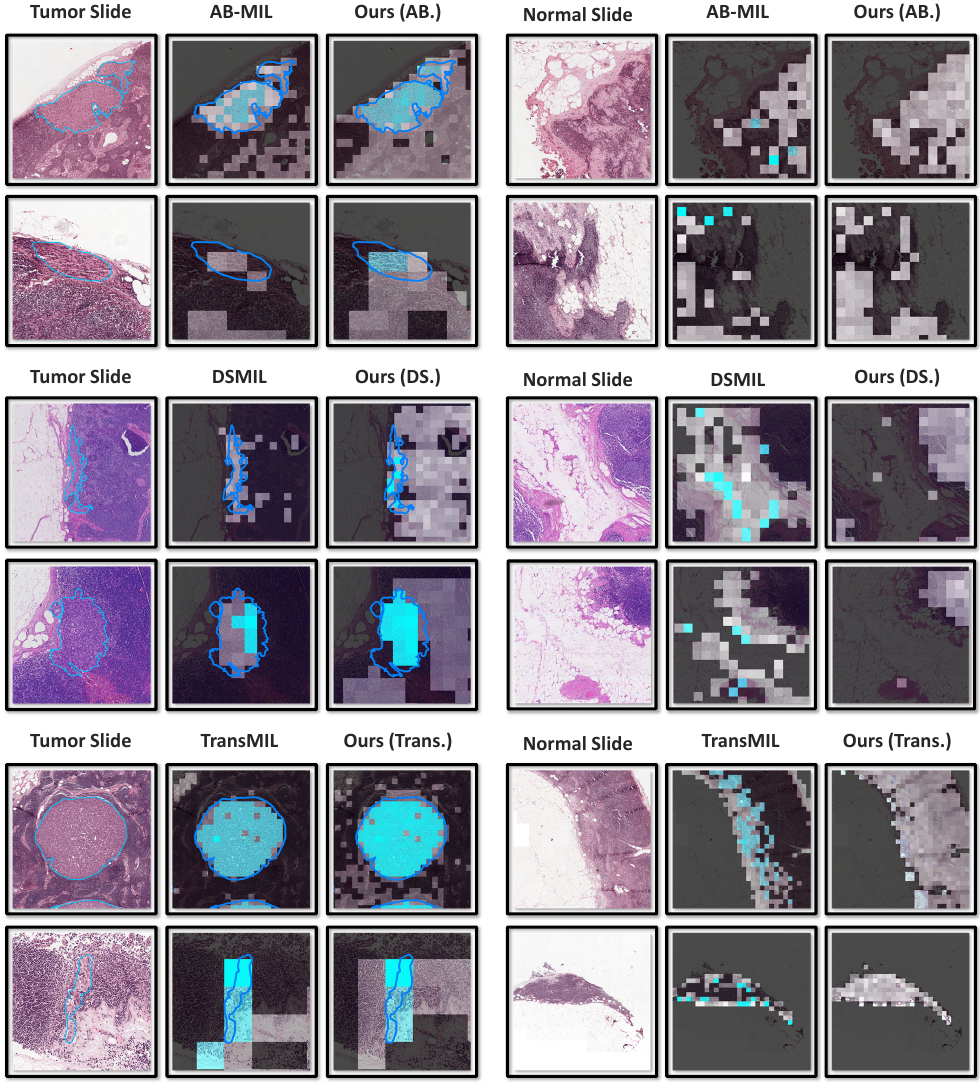}
    \caption{Patch visualization produced by baselines and MHIM-v2 on CAMELYON. The \textcolor{blue}{blue} lines outline the tumor regions. Brighter patches indicate higher attention scores. The \textcolor{cyan}{cyan} colors represent high probabilities of tumor presence at the corresponding locations. In the cyan patches, brightness reflects the confidence level. Ideally, the bright \textcolor{cyan}{cyan} patches should cover only the area within the \textcolor{blue}{blue} lines. We demonstrate that focusing solely on more salient regions reduces the generalization ability of baseline models. MHIM-v2 corrects the attention regions of baselines, making accurate and robust judgments from a pathologist's perspective.} 
    \label{fig:vis}
\vspace{-0.3cm}
\end{figure*}

\subsubsection{More on Masking Strategy}
\noindent \textbf{Error Masking Mitigation.~}
MHIM faces a major challenge: it may mask all key information and turn into ``error instance mining".
To further minimize this risk, we employ additional strategies, including a \textbf{decaying mask ratio (Decay) and a randomly high-score masking policy (RHSM)}. While the lack of instance-level supervision prevents the complete elimination of this risk, our quantitative and qualitative results demonstrate its effective mitigation. The ablation study in Table~\ref{tab:decay} shows that removing the decay and randomly masking strategies (RHSM) leads to a consistent performance drop.

\noindent\textbf{Discussion on Randomly Masking.~}
To isolate the impact of our hard instance mining strategy from simple random masking, we conducted an ablation study. In Figure~\ref{fig:hsm}, we include results for an RSM-only baseline (i.e., applying only random masking at various ratios) and mark the performance of our full MHIM framework for direct comparison. While RSM alone improves performance over the standard baseline by reducing data redundancy and increasing training diversity, the results clearly show a significant performance gap between the RSM-only approach and our full framework. This performance gap underscores the critical contributions of our proposed components: RHSM, CL, and GRN, in effectively guiding the model to focus on the most informative instances.

\subsubsection{Discussion on Hyperparameters}
We analyze the key hyperparameters in MHIM-v2. Specifically, the two mask ratios influence the quality of hard instances, while the scaling factor balances the impact of self-supervised and fully supervised labels during model training. Figure~\ref{fig:hyper-para} reveals consistent performance trends across different offline features, demonstrating the generalizability. Moreover, the framework exhibits relatively stable performance. Although the CAMELYON dataset shows greater performance variation due to limited data, the framework maintains satisfactory stability in the NSCLC dataset as the sample size increases. Notably, different baselines exhibit varying sensitivity to hyperparameters: AB-MIL, characterized by stronger instance sparsity, typically requires a lower mask ratio, while TransMIL demands a higher ratio.

\begin{figure}[t]
\centering
    \includegraphics[width=0.9\linewidth]{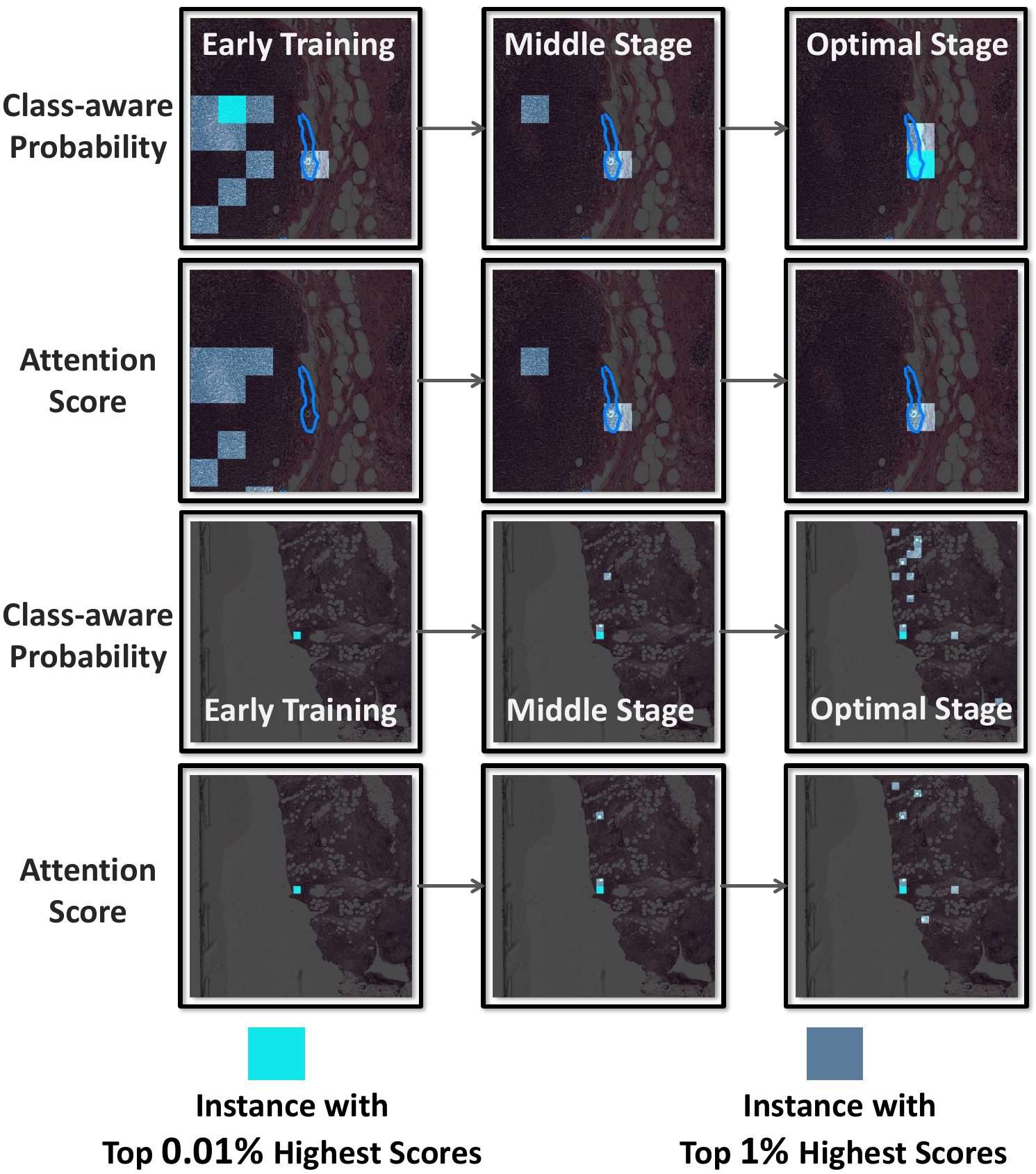}
    \caption{Visualization of easy-to-classify instances based on different strategies during the training process by the teacher model. As training progresses, the discriminability of teacher model gradually improves, and the advantage of class-aware instance probability over class-agnostic attention score becomes increasingly significant. This advantage is also shown in normal slides (bottom part of figure), where the evaluation distribution should be more uniform.} 
    \label{fig:vis_ds}
\end{figure}

\subsection{Visualization} 
\subsubsection{Diagnosis from Pathologist's view}
To intuitively understand the effect of masked hard instance mining, we visualize the attention scores (bright patches) and \textcolor{cyan}{tumor probabilities} (\textcolor{cyan}{cyan} patches) of patches produced by baselines and MHIM-v2, as illustrated in Figure~\ref{fig:vis}. MHIM-v2 employs AB-MIL (AB.), DSMIL (DS.), and TransMIL (Trans.) as its baseline models. We note that attention scores only indicate the regions of interest of models and are insufficient to reflect tumor probabilities~\cite{zhang2022dtfd,li2021dual}. Therefore, we additionally visualize tumor probabilities to better understand masked hard instance mining.

First, as shown in Figure~\ref{fig:vis}, we find that different attention regions determine the quality of discrimination. The baseline models focus only on tumor areas, leading to missing most of them. Expanding the view to include some "irrelevant areas" enables MHIM-v2 to make more complete judgments (rows \textcolor{dino_text}{1} and \textcolor{dino_text}{4} on the left). More importantly, hard instance mining enables the model to concentrate on the stained lymph nodes, thereby reducing the impact of noise in discrimination (rows \textcolor{dino_text}{3} and \textcolor{dino_text}{4} on the right). Clinically, lightly stained regions primarily consist of adipose tissue; any tumor tissues present are likely due to staining artifacts or sample contamination, leading to potential false positives. 
Thus, the model should avoid focusing on these areas, as they typically lack substantial tumor presence, and paying attention to these regions would suggest a misalignment in the model's target focus. 
In lymph node examinations, the primary diagnostic value lies in the lymph nodes themselves, with the lymph node capsule being of secondary importance. When examining lymph node metastasis under a microscope, pathologists typically start by inspecting along the capsule, as cancerous tissues generally first affect the marginal sinus near the capsule before infiltrating deeper. The MHIM-v2 model corrects the attention region bias found in baseline models, thereby enabling high-quality diagnoses from a pathologist’s perspective.

Moreover, baseline models often assign high tumor probabilities to patches in non-tumor areas. We attribute this phenomenon to the low generalization capability of conventional attention-based MIL models, which tend to focus only on salient regions during training. In contrast, MHIM-v2 trained with hard instances shows much better generalization ability than the baseline models for noise robustness (rows \textcolor{dino_text}{2} and \textcolor{dino_text}{6} on the right) and for precise detection of challenging subtle tumor areas (rows \textcolor{dino_text}{2} and \textcolor{dino_text}{3} on the left). This phenomenon demonstrates how hard instances provide more useful information to help the model make more accurate, and robust judgments.

\subsubsection{How MHIM Facilitates Training}
The assessment of easy-to-classify instances by the teacher model directly affects the quality of the mined hard instances, thereby influencing the training of the MHIM framework. Therefore, we visualize the teacher model's assessment of easy-to-classify instances at different training stages in Figure~\ref{fig:vis_ds} and qualitatively analyze the impact of the proposed class-aware instance probability.

From the upper part of the figure, we observe a gradual improvement in the discriminative ability of the teacher model, demonstrating the effectiveness of iterative optimization. More importantly, for tumor slides, the initial assessments are relatively random, making it difficult to accurately locate highly challenging tumor areas. In this scenario, the quality of the mined hard instances is poor. As training progresses, the teacher model can accurately evaluate easy-to-classify instances, leading to more challenging hard instances and better training outcomes. This clearly illustrates the necessity of the RHSM strategy, which prevents the teacher model from masking all discriminative instances in the later stages of training and providing incorrect instance sequences. Conversely, the teacher exhibits opposite characteristics on normal slides in the lower part of the figure. In the early stages of training, the teacher model is easily influenced by a small amount of noise, resulting in sparse assessment distributions and misjudgments. However, subsequent visualizations demonstrate that MHIM training enables the teacher to provide ideal, uniform assessments on normal slides. Additionally, the teacher can remove extreme interference for the student in the early stages of training, preventing the student model from falling into the same dilemma and facilitating the iterative optimization of MHIM.

Although the teacher model based on attention scores shows a trend similar to the proposed class-aware instance probability, we find that instance probabilities are more accurate and exhibit a more ideal distribution. This advantage is evident in both types of slides and becomes increasingly pronounced as training progresses.

\section{Conclusion}

This paper rethinks the impact of salient instances on MIL-based CPath algorithms. We demonstrate that attention-based MIL, which excessively prioritize salient instances, harm the generalization ability of the model.
To address this issue, we propose the masked hard instance mining that masks out easy-to-classify patches and encourages the model to focus on informative and clinically valuable regions for better learning.
Through qualitative analysis, we show that the proposed strategy effectively alleviates the underfitting problem of general attention-based MIL models to hard instances.
We also develop the MHIM-v2 framework, leveraging a momentum teacher, a global recycle network, class-aware instance probability, and consistency loss to further enhance hard instance mining.
Our experimental results demonstrate the superiority and generality of the MHIM-v2 framework compared to other SOTA methods and the earlier conference version.

\section*{Acknowledgements}
This work was supported in part by the National Natural Science Foundation of China under Grant 62176030, in part by the Key Projects of the Special Program for Technological Innovation and Applied Development in Chongqing Municipality under Grant CSTB2023TIAD-KPX0060.

\section*{Data Availability Statement}
\textbf{Data publicly available in a repository:}

The CAMELYON dataset is available at \url{https://camelyon17.grand-challenge.org/}. 

All TCGA datasets can be found at \url{https://portal.gdc.cancer.gov/}.

All CPTAC datasets can be found at \url{https://pdc.cancer.gov/}.

\bibliography{sn-bibliography}

\end{document}